\documentclass[conference]{IEEEtran}

\usepackage[utf8]{inputenc} 
\usepackage[T1]{fontenc}    
\usepackage{hyperref}       
\usepackage{url}            
\usepackage{booktabs}       
\usepackage{amsfonts}       
\usepackage{nicefrac}       
\usepackage{microtype}      
\usepackage{graphicx}
\usepackage{subfig}
\usepackage{hanging}
\usepackage{siunitx}
\usepackage[square,numbers]{natbib}
\title{Noise Sensitivity-Based Energy Efficient and Robust Adversary Detection in Neural Networks}

\IEEEoverridecommandlockouts
\usepackage{amsmath,amssymb,amsfonts}
\usepackage{algorithmic}
\usepackage{graphicx}
\usepackage{textcomp}
\usepackage{xcolor}
\def\BibTeX{{\rm B\kern-.05em{\sc i\kern-.025em b}\kern-.08em
    T\kern-.1667em\lower.7ex\hbox{E}\kern-.125emX}}

\begin{document}

\author{\IEEEauthorblockN{Rachel Sterneck}
\IEEEauthorblockA{\textit{Department of Electrical Engineering} \\
\textit{Yale University}\\
New Haven, USA \\
rachel.sterneck@yale.edu}
\and
\IEEEauthorblockN{Abhishek Moitra}
\IEEEauthorblockA{\textit{Department of Electrical Engineering} \\
\textit{Yale University}\\
New Haven, USA \\
abhishek.moitra@yale.edu}
\and
\IEEEauthorblockN{Priyadarshini Panda}
\IEEEauthorblockA{\textit{Department of Electrical Engineering} \\
\textit{Yale University}\\
New Haven, USA \\
priya.panda@yale.edu}
}

\maketitle

\begin{abstract}
Neural networks have achieved remarkable performance in computer vision, however they are vulnerable to adversarial examples. Adversarial examples are inputs that have been carefully perturbed to fool classifier networks, while appearing unchanged to humans. Based on prior works on detecting adversaries, we propose a structured methodology of augmenting a deep neural network (DNN) with a detector subnetwork. We use \textit{Adversarial Noise Sensitivity} (ANS), a novel metric for measuring the adversarial gradient contribution of different intermediate layers of a network. Based on the ANS value, we append a detector to the most sensitive layer. In prior works, more complex detectors were added to a DNN, increasing the inference computational cost of the model. In contrast, our structured and strategic addition of a detector to a DNN reduces the complexity of the model while making the overall network adversarially resilient. Through comprehensive white-box and black-box experiments on MNIST, CIFAR-10, and CIFAR-100, we show that our method improves state-of-the-art detector robustness against adversarial examples. Furthermore, we validate the energy efficiency of our proposed adversarial detection methodology through an extensive energy analysis on various hardware scalable CMOS accelerator platforms. We also demonstrate the effects of quantization on our detector-appended networks.
\end{abstract}

\section{Introduction}
Deep neural networks have rapidly accelerated the field of machine learning, demonstrating state-of-the-art performance on a variety of difficult tasks, including  computer vision \cite{he2016compvision},  speech recognition \cite{amodei2016speech}, and biomedical image analysis \cite{shen2017bio}. Although neural networks have great potential to be used in real-world vision tasks, their vulnerability to \textit{adversarial attacks} is a bottleneck that must be addressed to ensure that machine learning applications are safe and reliable. Recent research demonstrates that high performing models are vulnerable to small, calculated perturbations applied to images that are capable of fooling a network into misclassifying an input, yet are often imperceptible to humans \cite{biggio2013evasion} \cite{szegedy2014nn}  \cite{goodfellow2014badv}.  In a black-box attack scenario, hackers can create adversarial examples without knowledge of a target model's parameters by using another network to generate transferable attacks \cite{papernot2016ablackbox}. Furthermore, physical-world adversarial attacks have also fooled classification networks, including printed adversarial examples recaptured with a cell phone camera \cite{kurakin2016adversarial} and stop signs modified with tape perturbations mimicking graffiti \cite{eykholt2018robust}. These examples demonstrate the high-risk nature of adversarial examples, as well as the need to implement defenses against such attacks. 

In this paper, we introduce a method for detecting adversarial examples that utilizes the structure of the Convolutional Neural Network (CNN). We use \textit{Adversarial Noise Sensitivity} (ANS) to identify the layers that are most vulnerable to adversarial examples \cite{panda2020quanos}. The idea is to use ANS to determine which layer should be augmented with a detector, improving the robustness of the network in a compute-efficient manner. The detector is a binary classifier trained on intermediate layer activations to distinguish adversarial examples from clean inputs. After identifying a vulnerable layer of the trained CNN to augment with a detector, we use the intermediate activations from that layer to train the detector. Our results on MNIST, CIFAR-10, and CIFAR-100 empirically show that our method achieves state-of-the-art detection robustness against various adversarial attacks.

To facilitate compatibility with practical hardware accelerators, we quantize the proposed detector-augmented CNN architecture to smaller bit-widths per layer and compare to a 16-bit baseline model. We consider quantized models because network quantization reduces the overall energy consumption of the model by decreasing the number of bits used to represent network parameters, including activations and weights. Given the large number of parameters in a CNN, quantization is a useful method for improving the energy efficiency of accessing and storing DNN parameters without compromising accuracy. We observe that the detector maintains high accuracy, even at 1-bit quantization. The quantized CNN with detector network is implemented on a precision scalable hardware accelerator that can support hardware scalability paradigms such as Data Gating (DG) and Dynamic Voltage-Accuracy and Frequency Scaling (DVAFS) \cite{moonsdg} \cite{moonsdvafs}. 

We conduct energy analyses to show that the quantized model consumes considerably less energy than the non-quantized 16-bit CNN and detector model. Furthermore, we demonstrate the energy efficiency of our early adversarial detection scheme by comparing the Multiply and Accumulate (MAC) computation and memory access energies of the standalone CNN network to the quantized detector-appended CNN architecture under various scenarios. We find that the quantized detector-appended CNN network yields higher energy efficiency than the standalone network for different concentrations of adversarial examples in the test dataset.

To summarize, we make the following contributions in this paper:
\begin{enumerate}
    \item We improve state-of-the-art detection-based systems for identifying adversarial examples on the CIFAR-10 and CIFAR-100 datasets based on ANS, which allows for a structured addition of detectors to CNNs.
    \item We demonstrate the energy efficiency of our proposed detector-appended CNN architecture with respect to a standalone CNN network by performing MAC and memory access energy estimation on a precision scalable hardware accelerator with support for DG and DVAFS computation paradigms.
\end{enumerate}

\section{Background}
\subsection{Generating Adversaries}
The architecture, parameters, and gradients of a CNN are used to generate adversarial examples. Over the last few years, several methods for creating adversarial examples have been developed, and here we provide an overview of two widely used methods: \textit{Fast Gradient Sign Method} (FGSM) and \textit{Projected Gradient Descent} (PGD).  

\textbf{\textit{Fast gradient sign method}} is a simple method which crafts adversarial examples by linearizing a trained model's loss function ($L$, e.g. cross-entropy loss) with respect to the input ($X$) \cite{goodfellow2014badv}:
\begin{equation}
    x_{adv}=x+\varepsilon sgn(\nabla_xL(\theta, x, y))
\end{equation}
Here, $y$ is the true class label for the input image $x$, $\theta$ denotes the model parameters (weights, biases, etc.) $\varepsilon$ is the $L_{\infty}$-constrained attack strength, and $sgn$ is the sign of the gradient of the loss function. FGSM is not as effective as other algorithms for generating adversaries; however, in our work it provides meaningful insights related to ANS and attack generalizability. 

\textbf{\textit{Projected gradient descent}} is regarded as the most effective adversarial attack generated from a network's local first order information, approximately optimizing the perturbations \cite{madry2017towards}. PGD is an iterative attack that recalculates the gradient and adds perturbations in each iteration:
\begin{equation}
    x_{adv}^{t+1}=\Pi_{x+S}(x^{t}+\alpha sgn(\nabla_xL(\theta, x, y)))
\end{equation}
The PGD algorithm is similar to FGSM, however  $x_{adv}^{0}$ is initialized by adding some random perturbation to the input image,  $x_{adv}^{t+1}$ represents the adversarial example of the current iteration, $S$ denotes the set of images in the dataset, and $\alpha$ is the step size in the direction of the sign of the gradient.  
A key point here is that gradient propagation is a crucial step in adversarial input generation. This implies that the adversarial gradient contribution to the net perturbation ($sgn(\nabla_xL(\theta, x, y))$) from different layers can vary depending upon the learned activations, which in turn provides the motivation for evaluating the ANS per layer (described more in Section \ref{methodology-ans}). 
\subsection{Hardware Realization}
\subsubsection{Quantization}
In light of the enormous compute power required by deep learning, quantization is a method of reducing the bit-precision of neural network parameters  in order to perform faster and energy efficient computations on large scale hardware accelerators. Recent works have also shown that quantization prevents DNNs from overfitting, and therefore has useful implications, including improved robustness and energy efficiency, without accuracy degradation \cite{panda2020quanos}. Quantization thus serves as a reasonable hardware-centric solution that can be used to realize practical DNN accelerators. Typically, the most common ways of implementing quantization are by using 1) homogeneous bit-precision throughout all layers in the network \cite{wu2016quantized} or 2) mixed-precision, i.e. different layers having different data precision \cite{panda2020quanos}.

\subsubsection{Hardware Scaling Paradigms}
In terms of DNN accelerator design, parameters such as data and instruction parallelism, voltage of operation, and clock frequency are crucial factors that affect the accelerator performance and energy consumption. One might choose to set the parameters to standard values, which are referred to as the \textit{nominal value}, or use a value lower than the \textit{nominal value} called the \textit{scaled value}. In this section, we briefly discuss two hardware scaling paradigms: \textit{Data Gating} (DG) and \textit{Dynamic Voltage-Accuracy and Frequency Scaling} (DVAFS) \cite{moonsdg} \cite{moonsdvafs}.  

\begin{figure}[!htb]%
    \centering
    \qquad{{\includegraphics[width=8cm]{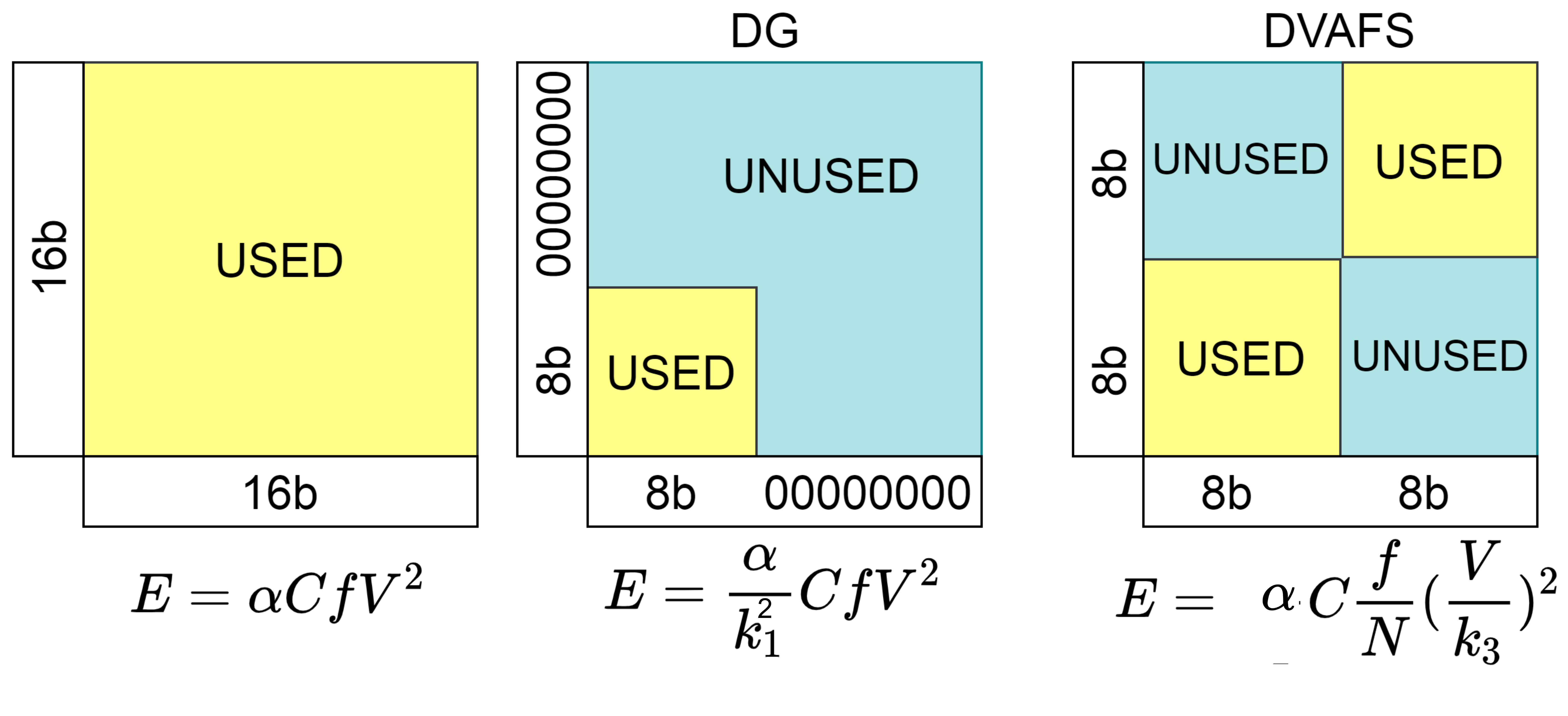} }}%
    \caption{Comparison of DG and DVAFS hardware scalibility paradigms with standard 16-bit MAC computation.}%
    \label{dg_dvafs_hw}%
\end{figure}

\textbf{\textit{Data gating}} is a method where the data precision during MAC computation is scaled from the nominal 16-bit precision to a smaller value (i.e. 8-bit) to shorten the critical path of the circuit (which implies faster calculations) such that the supply voltage could be scaled to a lower value. The DG device in Fig. \ref{dg_dvafs_hw} demonstrates that scaling down the voltage \textit{V} by a factor of $k_{1}$ leads to a $k_{1}^2$ times reduction in the consumed energy compared to the standard 16-bit MAC operation. It must be noted that the MAC operation occurs between the Most Significant Bits (MSBs) of the input and synaptic weights while the Least Significant Bits (LSBs) are gated and therefore are not used in the computation, which saves energy.

\textbf{\textit{Dynamic voltage-accuracy and frequency scaling}} is a paradigm shown in Fig. \ref{dg_dvafs_hw} where data-parallelism is implemented at the sub-word level, unlike data-scaling in DG. A sub-word refers to a scaled factor of the maximum word length, for example 2-bit, 4-bit and 8-bit. To illustrate sub-word level parallelism, consider the bit-width of a layer to be 8-bit (which is equal to one of the 2 sub-words mentioned above). In a 16-bit MAC unit, $2\times$8-bit (and $4\times$4-bit in the case of 4-bit sub-word) MAC operations can be performed in parallel. This implies that at the same throughput requirement, the frequency of operation can be lowered by \textit{N}=$2\times$ for 8-bit data and \textit{N}=$4\times$ for 4-bit data. This lowers the energy consumption by a factor \textit{N}. Furthermore, lowering the voltage V by factor $k_3$ reduces the energy consumption by a factor of $k_3^2$. Also, it can be observed that at sub-word bit-widths, DVAFS utilizes the MAC unit more efficiently than DG \cite{moons2020dvafs, moons2016energy}.

While DG and DVAFS paradigms improve the energy efficiency of DNN accelerators, the hardware design loses generalization across different bit-precisions. Thus, if the data precision changes, the hardware architecture needs to be re-designed to handle the new data-precision. However, in this work, we avoid this problem by specifically using two data precisions: 8-bit precision and 12-bit precision. 
\subsection{Related Work}
\subsubsection{\textbf{Adversarial Detection}}
Many detection methods have emerged to counter adversarial examples, ranging from augmenting the dataset to modifying the architecture of the network. For example, Goodfellow et al. \cite{goodfellow2014badv} train the model on both clean and adversarial examples. Likewise, Grosse et al. \cite{grosse2017detection} use an additional $K+1$-th class for identifying adversarial examples, and Gong et al. \cite{gong2017adversarial} train a  binary classifier to discriminate between real and adversarial inputs as a preprocessing step. Similarly, Yin et al. \cite{yin2020gat} develop a method involving input space partitioning and training binary classifiers in subspaces. Most relevant to our work, Metzen et al. \cite{metzen2017detecting} append a binary classifier between convolutional layers to detect intermediate feature representations of an adversary. We consider this to be a more robust approach for detection-based methods because it requires an attacker to not only perturb the input images, but also to access and modify the intermediate activations in order to attack the whole system.

Additionally, Bhagoji et al. \cite{bhagoji2018enhancing}
reduce the dimensionality of the input images fed to the classification network and train a fully-connected
neural network on the smaller inputs. Li and Li \cite{li2017adversarial} build a cascade classifier where each classifier is
implemented as a linear SVM acting on the PCA of inner convolutional layers of the classification
network. Feinman et al. \cite{feinman2017detecting} add a classifier to the final hidden layer of a CNN using a kernel density estimate method to detect the points lying far from the data manifolds in the last hidden layer. However, Carlini and Wagner \cite{carlini2017adversarial} show that each of these defense methods can be evaded by an adversary targeting that specific defense, i.e. by a white-box adversary. 

\subsubsection{\textbf{Hardware approaches}}
Beyond pure algorithmic approaches for minimizing the effects of adversarial attacks, recent works have attempted to solve the problem of adversarial robustness using hardware-algorithm co-design approaches. The authors in  \cite{panda2019discretization} show that input data and parameter quantization significantly contribute to improvements in adversarial robustness. The paper also shows that \textit{Binary Neural Networks} (BNNs) exhibit higher adversarial robustness than multi-bit precision networks, and the adversarial robustness of BNNs can be further improved by input data quantization. The work on Defensive Quantization (DQ) employs data quantization homogeneously to each layer of the DNN in order to reduce the error magnification effect in deeper layers by ensuring that the Lipschitz constant value of the network \cite{cisse2017parseval} is less than or equal to 1 \cite{lin2019defensive}. Another prominent work is QUANOS \cite{panda2020quanos}, which uses mixed precision quantization for different layers in order to introduce robustness during training. This allows the mixed quantized models to perform better than baseline models under FGSM and PGD attacks.

The work by Bhattacharjee et al. \cite{bhattacharjee2020rethinking} discusses the benefits of crossbar non-idealities in light of adversarial robustness. This work shows that device non-idealities in analog crossbars can improve the adversarial robustness by approximately 10-20\%. Furthermore, the work on \textit{Conditional Deep Learning} (CDL) by Panda et al. \cite{panda2016conditional} shows that early detection of non-adversarial examples can be very useful in performing energy efficient inference. Here, small linear classifiers are strategically introduced at the end of selected layers to perform early inference. An early-exit approach prevents unnecessary propagation through CNN layers, which in turn saves energy. 

Additionally, there are other emerging classes of neural attacks prompting novel strategies for adversarial defense, including trojan (or backdoor) attacks. Unlike adversarial attacks, which only require input images to be perturbed by an adversary, neural trojan attacks involve injecting malicious behavior into the model that can be activated by special inputs called "triggers." 
Several techniques have been proposed to defend against trojan attacks: Xu et al. \cite{xu2019detecting} use a meta-classifier network to detect trojaned DNN model outputs and mitigate trojan attacks. Likewise, Gao et al. \cite{gao2019strip} use the entropy parameters of DNNs to detect trojaned inputs; a low DNN entropy value violates the input-dependence characteristics of benign models and suggests trojaned inputs. Tran et al. \cite{tran2018spectral} identify spectral signatures, a property found in backdoor attacks, which they use to detect poisoned inputs. 

In this work, however, we focus on adversarial attacks. We employ early detection of adversarial examples using a small neural network-based binary detector. The binary detector is strategically placed at the end of a specific CNN layer. This facilitates conditional propagation of activation values to deeper layers based on whether the input is classified as clean or adversarial. If the detector classifies the image as adversarial, then its propagation to the later layers is terminated, which prevents unnecessary computations. Furthermore, our proposed ANS-based adversarial detection provides a structured and computationally efficient approach for filtering adversaries, while also improving the detection rates of previous works. By appending the most sensitive layer, i.e. the layer with the highest contribution to adversarial perturbations, with a simple detector, we demonstrate that our method yields strong resilience even against dynamic white-box adversaries wherein both the detector and the network are attacked, as well as black-box attacks involving adversarial examples generated by a substitute model.
\section{Methodology} \label{methodology-section}
\subsection{Adversarial Noise Sensitivity (ANS)}  \label{methodology-ans}
\begin{figure}%
    \centering
    \subfloat[ANS for FGSM attacks of varying strength.]{{\includegraphics[width=8cm]{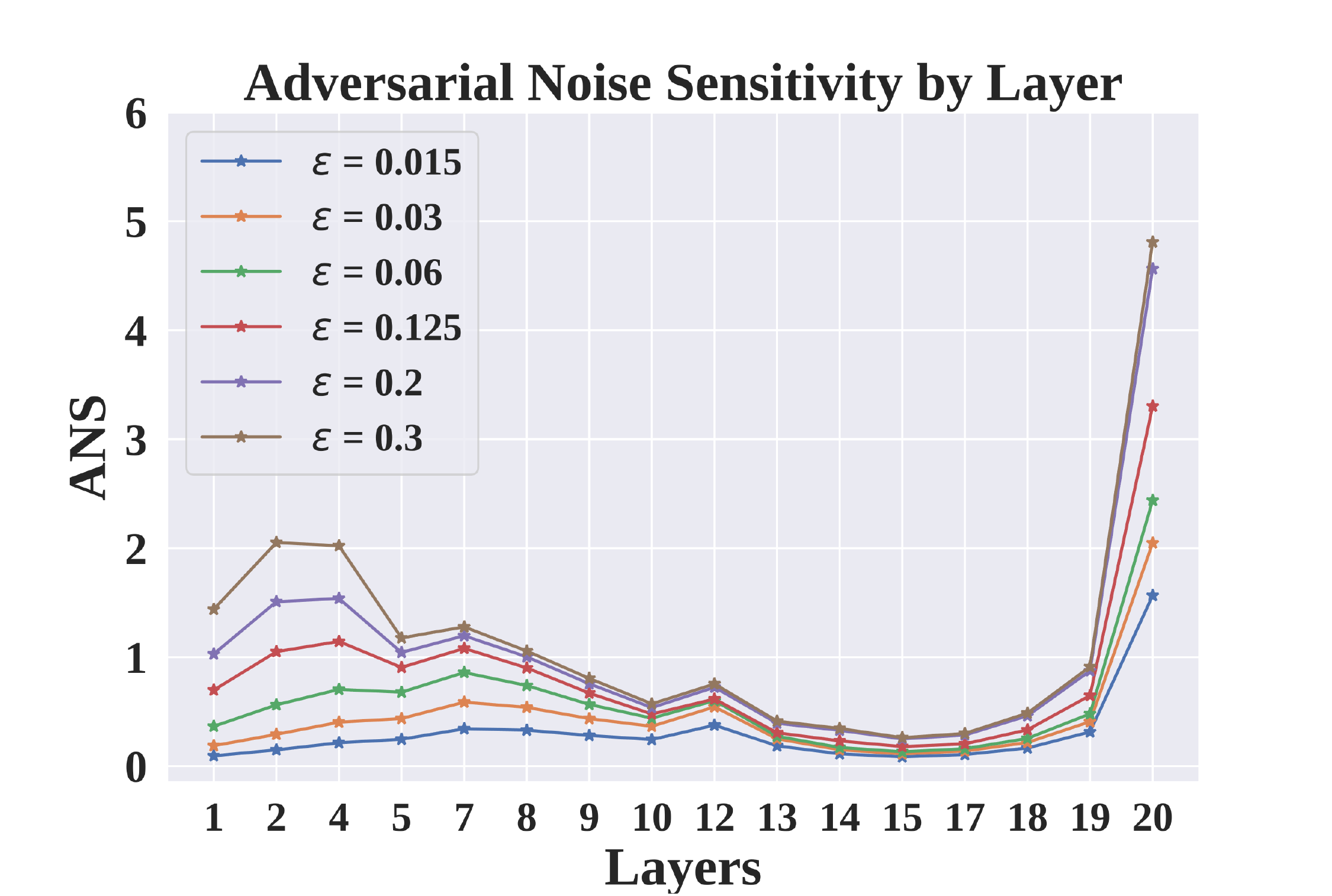}}}%
    \qquad
    \subfloat[ANS for PGD attacks of varying strength. ]{{\includegraphics[width=8cm]{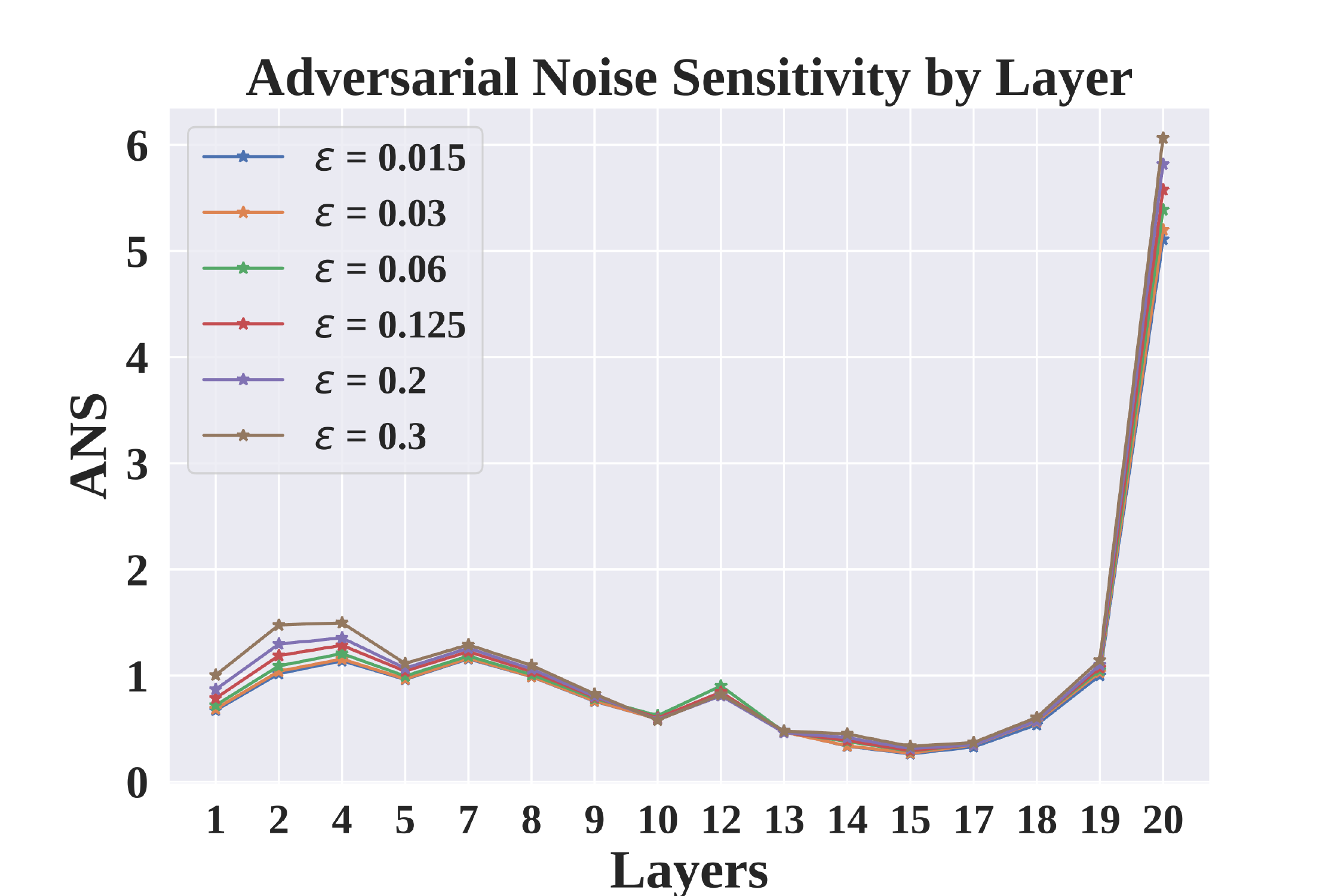}}}%
    \caption{ANS insights for VGG19 and CIFAR-10. The sensitivity trend is similar across different adversarial attacks, and layer 7 is identified as a good detector candidate for detectors trained on low-strength PGD attacks.}%
    \label{fig:ans-vgg19}%
\end{figure}

A novel outcome of this work is the application of a new form of noise stability for DNNs, as well as a method for using this metric to evaluate where in the network a detector should be added. ANS provides layer-wise sensitivities to adversarial inputs, and is computed as follows:
\begin{equation}
    ANS_l = \sqrt{(a_{adv}^{l} - a^{l})^2}
\end{equation}
Here, $a_{adv}^{l}$ and $a^{l}$ are activation values of layer $l$ when the adversarial input ($x_{adv}$) and the clean input ($x$), respectively, are passed through the network. High ANS implies more changes in activations, which can be attributed to high adversarial contribution by a layer. Fig. \ref{fig:ans-vgg19}(a),(b) show ANS values for different convolutional layers of a VGG19 model trained on CIFAR-10 data when exposed to FGSM and PGD attacks, respectively. The graphs demonstrate that ANS trends are generally consistent for different attack strengths and types. For example, we identify high ANS peaks at layers 4 and 7 in both Fig. \ref{fig:ans-vgg19}(a),(b), even for different FGSM and PGD attacks, which indicates that these layers are more vulnerable to adversarial attacks than layers with lower ANS values. 

\begin{figure}[!htb]%
    \centering
    \subfloat[Adversarial accuracy for different amounts of pruning for an FGSM attack with $\varepsilon=0.3$. ]{{\includegraphics[width=8cm]{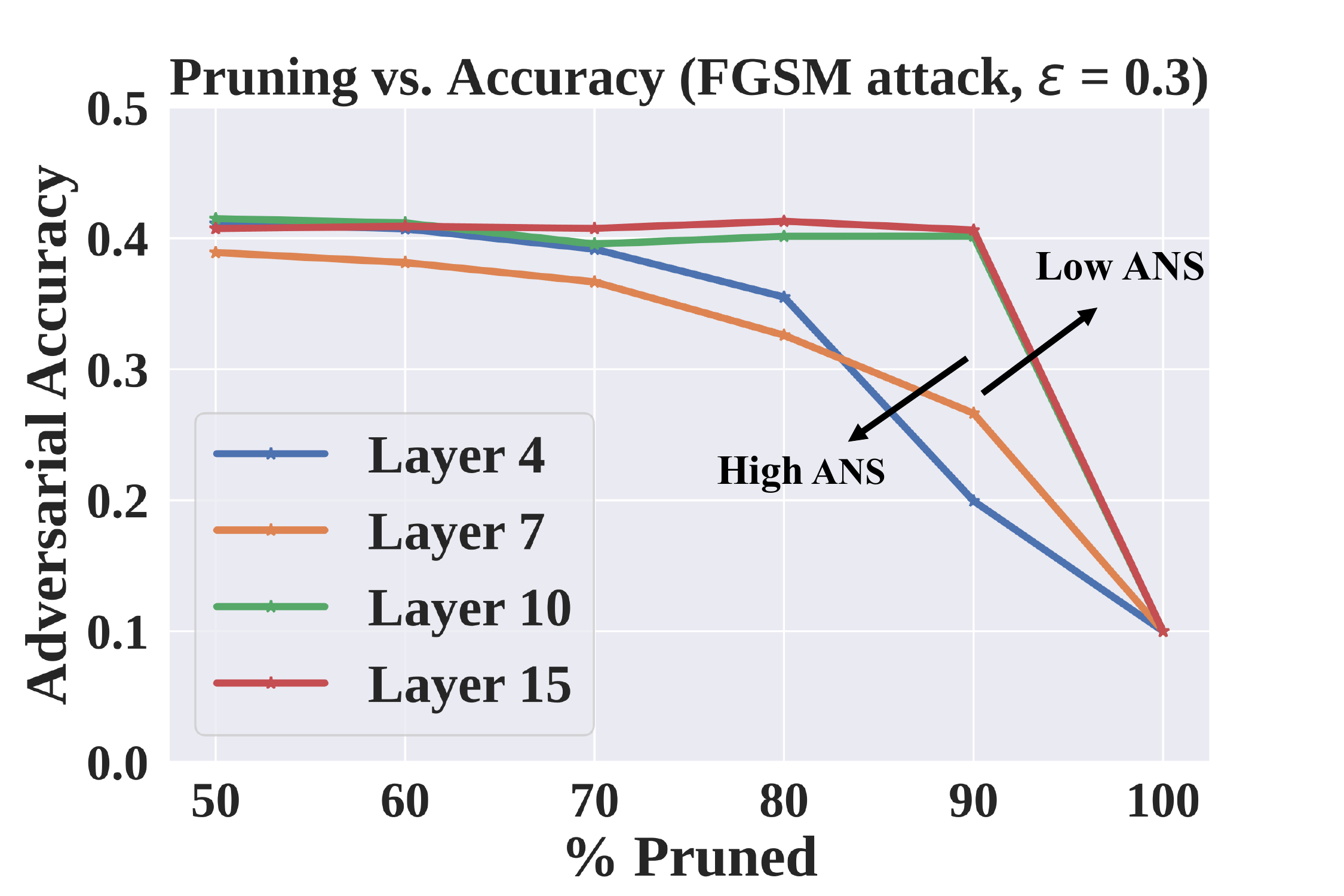} }}%
    \qquad
    \subfloat[Clean accuracy for different amounts of pruning. ]{{\includegraphics[width=8cm]{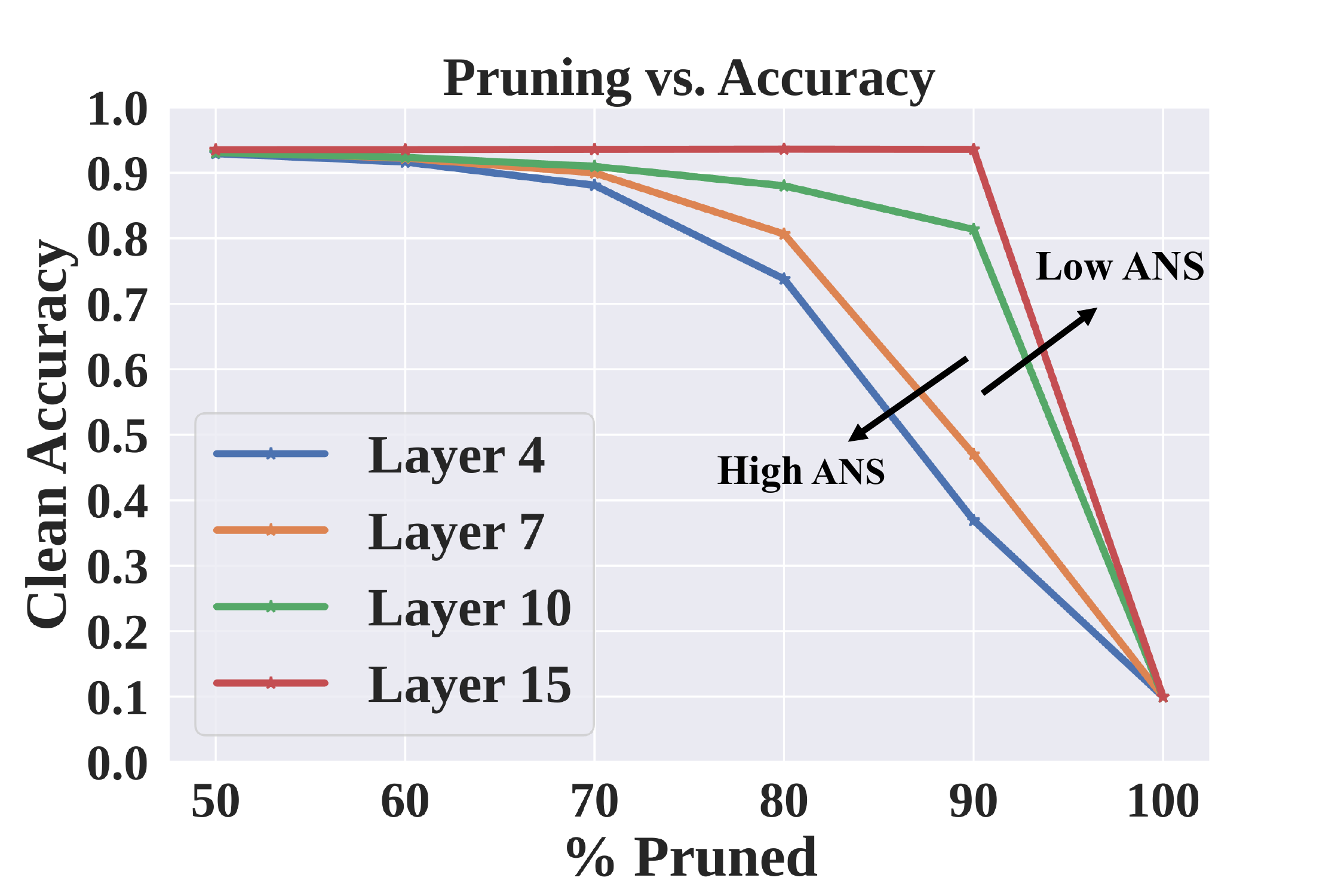} }}%
    \caption{Adversarial Noise Sensitivity (ANS) ablation tests. High ANS layers show a steeper decline in accuracy on CIFAR-10 when pruned.}%
    \label{fig:ans-ablation}%
\end{figure}

To further demonstrate the merit of ANS, we conducted an ablation study that measures the importance of a single direction (i.e. neuronal activations) to a network's computation by evaluating its performance as the direction is removed. Specifically, we measured the network's adversarial and clean test accuracies as we pruned the activations of intermediate layers' neurons, as shown in Fig. \ref{fig:ans-ablation}(a),(b). Fig.  \ref{fig:ans-ablation}(a) demonstrates that the decline in adversarial accuracy for high ANS layers is much steeper than that of low ANS layers. This implies that high ANS layers have more important directions and are more susceptible to adversarial attacks. Unless a low ANS layer is completely pruned, its adversarial accuracy is generally unaffected. Furthermore, Fig. \ref{fig:ans-ablation}(b) exhibits that this trend holds for ablation tests on the network with clean examples; again, the low ANS layers decline more gradually, compared to the steep decline of the high ANS layers. This may suggest that high ANS layers generally contribute more to the network overall, even when clean data is passed through. The consistencies between Fig. \ref{fig:ans-vgg19} and Fig. \ref{fig:ans-ablation} illustrate ANS as a powerful, yet simple, metric to analyze the contribution of each layer to the net adversarial perturbation during the gradient propagation. That being said, it's important to note that ANS is a heuristic value, and detector accuracy on a particular layer doesn't correspond one-to-one with its ANS value. For example, layers 18-20 in Fig. \ref{fig:ans-vgg19}(a),(b) have high ANS values, however an ablation test on these deeper layers reveals that their accuracies generally remain unchanged when the convolutional layer is pruned. We believe this happens because later layers predominantly learn high-level features and contribute less to the overall network accuracy than earlier layers that learn more primitive features do \cite{zeiler2014cnn}. 
\subsection{Detector} 
\begin{figure}[!htb]%
    \centering
    \qquad{{\hspace*{-1cm}\includegraphics[width=8cm]{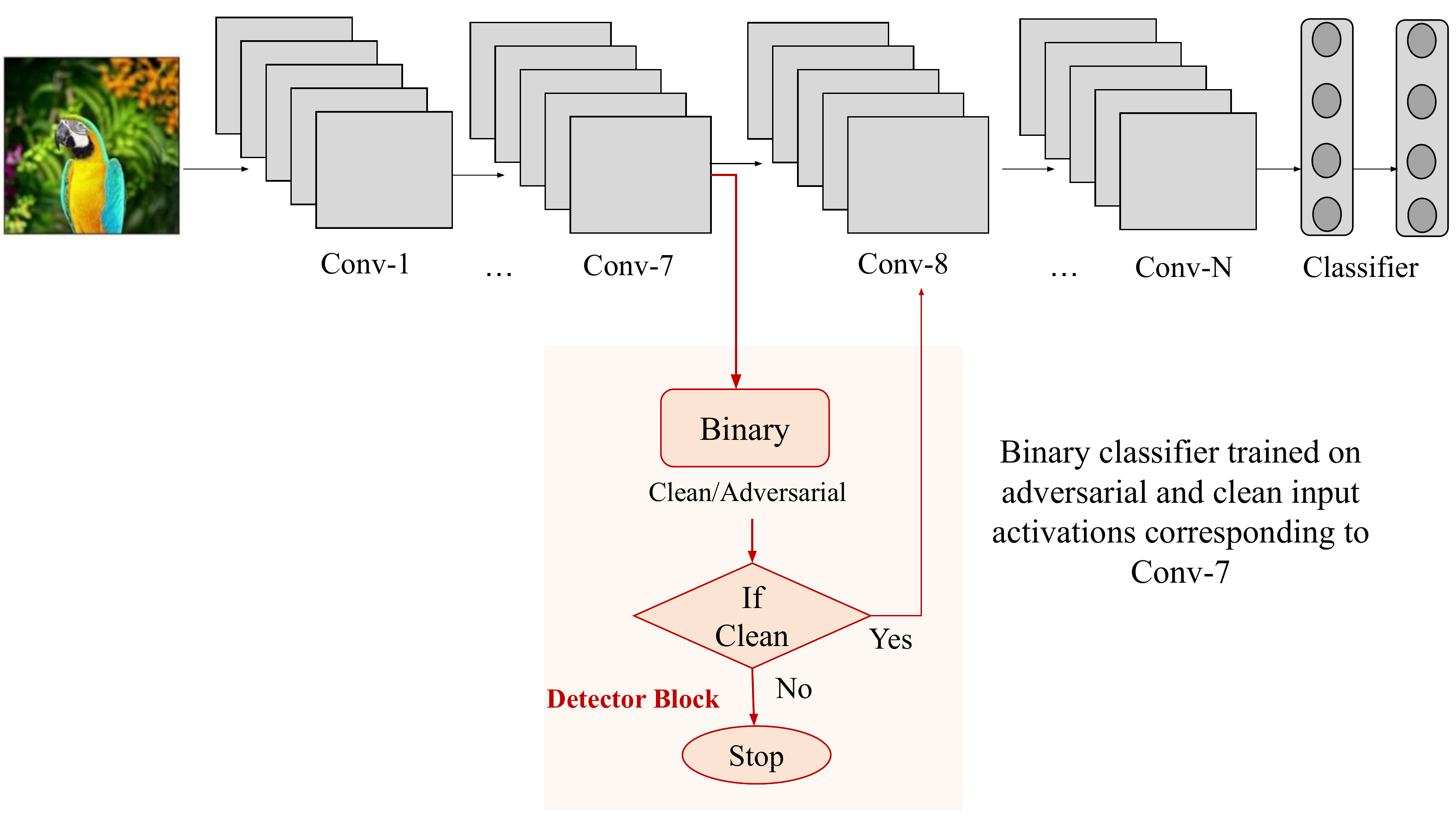} }}%
    \caption{The network architecture of a VGG19 augmented with a detector.}%
    \label{fig:detector-architecture}%
\end{figure}

We use ANS to determine which layer in the network must be followed by a detector. Note, we take a trained CNN and then train the detector separately on the intermediate activations of the trained CNN model. As shown in Fig. \ref{fig:detector-architecture}, we place the detector after a convolutional layer, excluding pooling and fully connected layers as candidates. The detector is a simple binary classifier with two fully connected layers. For our augmented LeNet and VGG19 networks used in Section \ref{experiments-section}, we create a training dataset of 51,200 samples, and for ResNet18, we use 25,600 samples; each testing dataset consists of 20,000 activations. Both our testing and training datasets include activations from the corresponding detector layer, and are equally comprised of clean and adversarial activations. Additionally, the detector is trained on a fraction of the original training dataset, thus this approach incurs lower training complexity. The detector is trained for a total of 30 epochs, with a learning rate of 0.03 for first 15 epochs and learning rate of 0.003 for the latter 15 epochs.

\section{Experiments}
\label{experiments-section}
\subsection{Datasets and Models}
In this work, we used PyTorch to train our models and conduct our experiments. For the MNIST \cite{lecun1998mnist} dataset evaluations, we used a standard LeNet \cite{lecun1989lenet} with $98.6\%$ accuracy trained for 30 epochs with a learning rate of 0.1. The MNIST dataset consists of 60,000 training images and 10,000 testing images. Additionally, for the CIFAR-10 and CIFAR-100 \cite{krizhevsky2009cifar}, 
datasets we used VGG19 \cite{simonyan2015vgg} and ResNet18 \cite{he2016compvision} models trained for 210 total epochs with a starting learning rate of 0.1 and a learning rate decay of 0.1 at step size 70. The VGG19 network achieves $93.7\%$ accuracy on CIFAR-10 and $70.3\%$ accuracy on CIFAR-100, and the ResNet18 network achieves $95.3\%$ accuracy on CIFAR-10 and $76.7\%$ accuracy on CIFAR-100. The CIFAR-10 and CIFAR-100 datasets consist of 50,000 training images and 10,000 testing images; given the large number of classes relative to the dataset size, CIFAR-100 is regarded as a more difficult classification task. Each model is trained with stochastic gradient descent as the learning algorithm. 
\subsection{Robustness Evaluation Metrics} \label{robustness-evaluation-metrics}
We use Area Under the Receiver Operating Characteristic Curve (AUC) to measure robustness against adversarial attacks. For detection-based approaches of identifying adversarial examples, AUC is a more comprehensive measure than accuracy because it reflects the true positive and false positive rates. The standard accuracy metric assumes that probability scores (softmax outputs) are properly calibrated by imposing a threshold score of 0.5 (i.e. scores above 0.5 are classified as positive and scores equal to or below 0.5 are classified as negative). On the contrary, AUC measures a classifier's ability to distinguish between classes, regardless of whether the probability scores are properly calibrated in the range [0, 1]. Intuitively, AUC is the probability that the classifier ranks a randomly chosen positive observation higher than a randomly chosen negative observation \cite{fawcett2005auc}. A detector with an AUC score of 1.0 is considered perfect, whereas one with an AUC score of 0 is entirely inaccurate. That being said, we find that AUC and accuracy generally follow each other, especially for white-box attacks. 
\subsection{Static White-Box Attack}
\label{static-whitebox-section}
In the static white-box threat model, we assume that the adversary attacks the input images with full knowledge of the CNN's parameters, but is unaware of the detection mechanism. Using the ANS heuristic, we comprehensively demonstrate detector robustness against static white-box attacks on the CIFAR-10, CIFAR-100, and MNIST datasets.

\begin{figure}[!htb]%
    \centering
     {{\includegraphics[width=6.5cm]{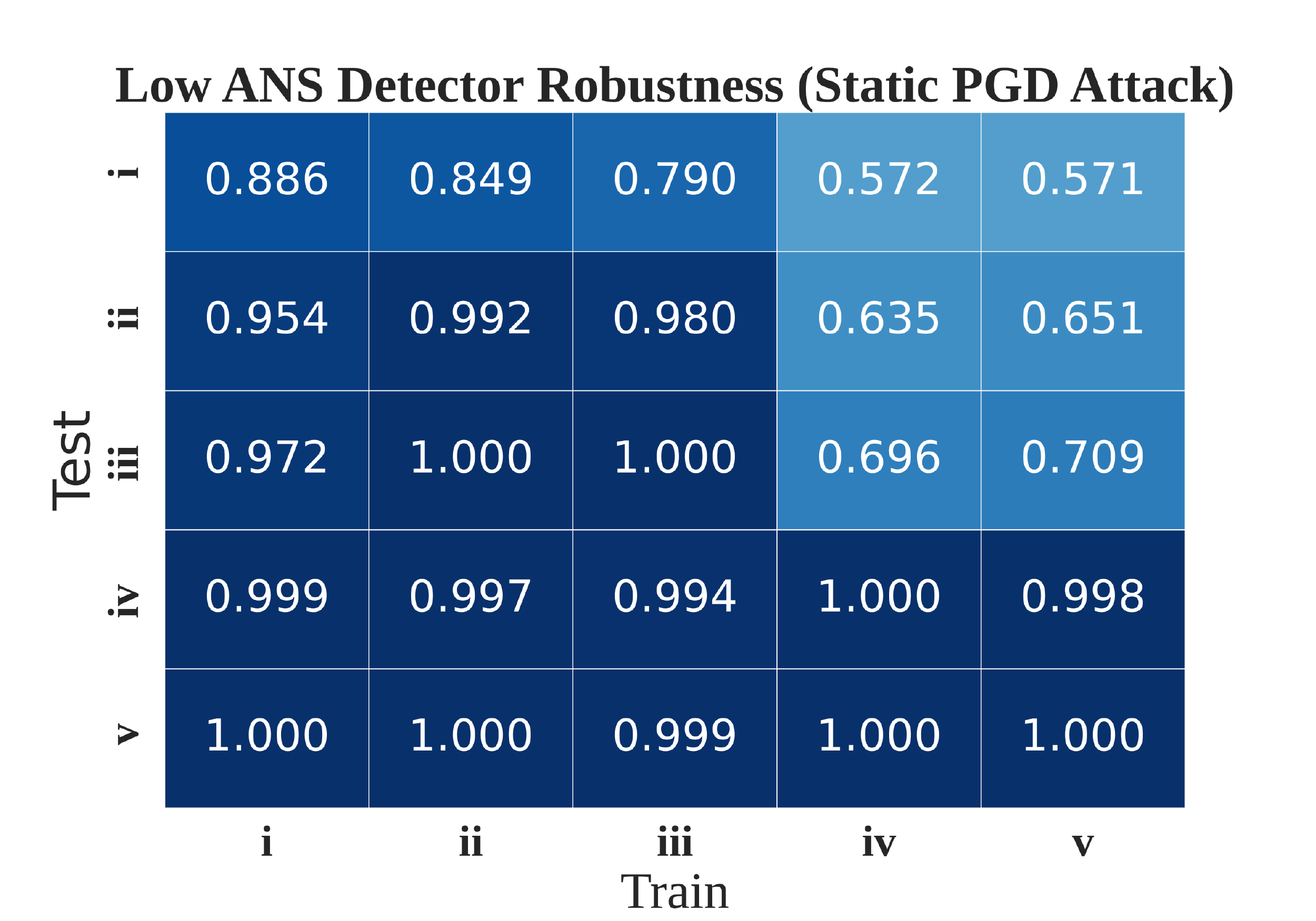} }}%
    \qquad{{\includegraphics[width=6.5cm]{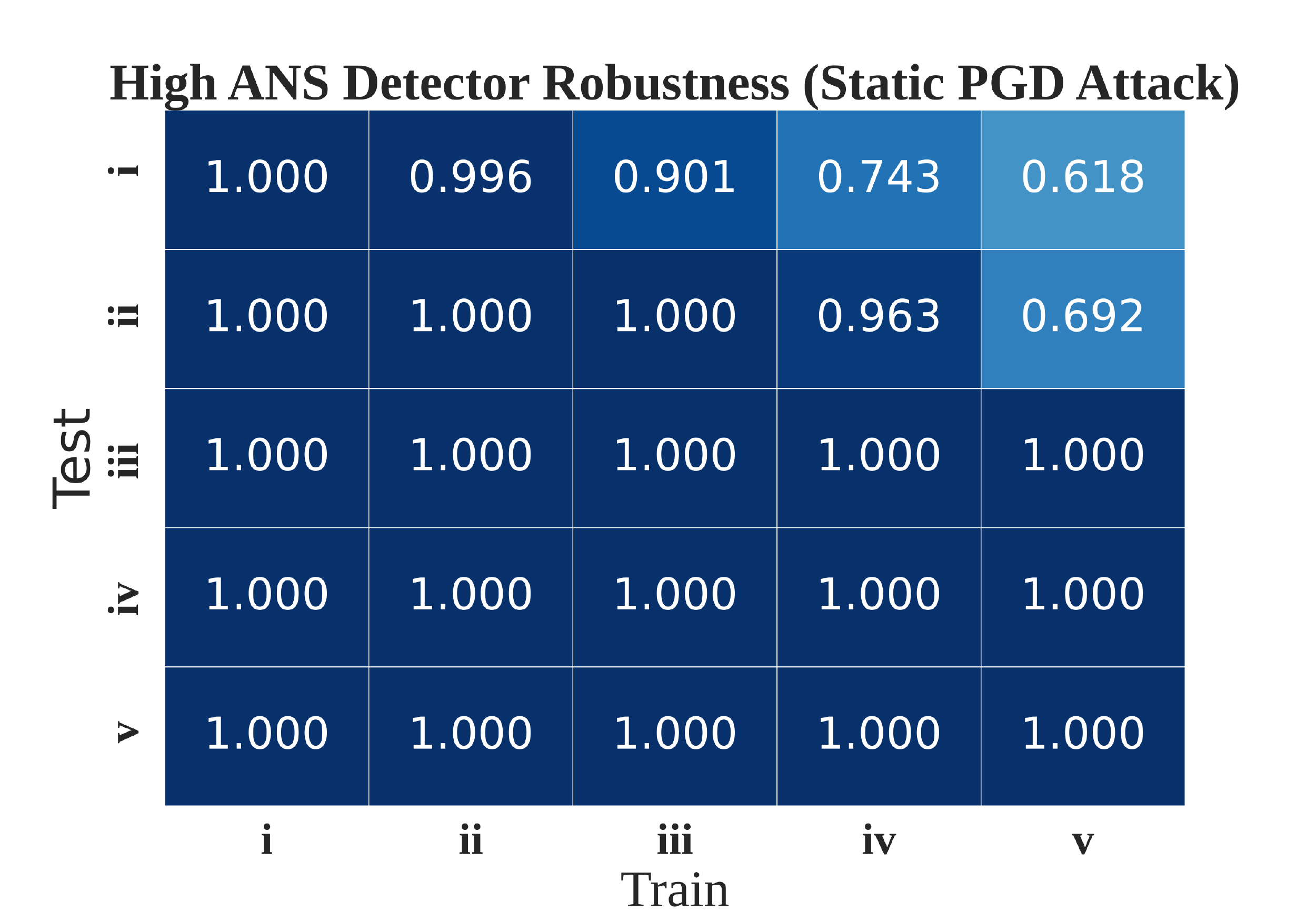} }}%
    \caption{AUC scores for static white-box PGD attacks on a VGG19 network augmented with a detector attached to low and high ANS layers. The adversarial examples are generated from CIFAR-10, and the attacks vary in strength, as referenced in Table \ref{static-cifar10-attack-params}.}%
    \label{fig:vgg-static-pgd}%
\end{figure}

For both the VGG19 and ResNet18 models, we train a binary classifier on activations from adversarial images that are generated from CIFAR-10 with the PGD attack algorithm. We train the detectors on five variations of PGD, with attack hyperparameters differing in the number of attack steps (\textit{n}), step size ($\alpha$), and attack strength ($\varepsilon$); the specific attack variations and details for static white-box attacks on CIFAR-10 are shown in Table \ref{static-cifar10-attack-params}. We provide results for VGG19 and ResNet18 neural networks augmented with a detector, demonstrating that ANS-based detectors are robust across different CNNs. Fig. \ref{fig:vgg-static-pgd} compares robustness of a VGG19 network augmented with a detector at a high ANS layer (layer 7) versus a low ANS layer (layer 15). Both detectors are trained and tested at various PGD attack strengths, and the results demonstrate that a detector placed after a higher ANS layer achieves greater robustness and generalizability between attack strengths, as measured by AUC. Furthermore, with an early-exit strategy for detecting adversarial examples, appending the detector at the end of layer 7 is more energy efficient than adding the detector after layer 15. We find that the VGG19 network augmented with a detector after layer 7 is most robust against static PGD attacks when trained on weaker attacks (i.e. attacks with smaller hyperparameter values \textit{n}, \textit{a}, \textit{e}): the detector trained on the PGD attack with $n=7$, $\alpha=0.007$, $\varepsilon=0.125$ achieves perfect detectability for a range of PGD attacks.

\begin{table}[hbt!]
  \caption{Training and testing hyperparameters for PGD static white-box attacks on CIFAR-10, which vary in the number of attack steps (\textit{n}), step size (\textit{a}), and attack strength (\textit{e}), as shown below.}
  \centering
  \resizebox{0.45\textwidth}{!}{\begin{tabular}{lllll}%
    \cmidrule(r){1-4}
         Attack Label    & Step Size ($n$) & Step Width ($\alpha$) & Epsilon ($\varepsilon$)      \\
    \midrule
    i &  7 & 0.007 & 0.125    \\
    ii &  20 & 0.007 & 0.125     \\
    iii & 100 & 0.007 & 0.125     \\
    iv &  40 & 0.5 & 8.0    \\
    v &  200 & 0.5 & 8.0   \\
    \bottomrule
    \label{static-cifar10-attack-params}%
  \end{tabular}}
\end{table}

\begin{figure}[!htb]%
    \centering{{\includegraphics[width=6.5cm]{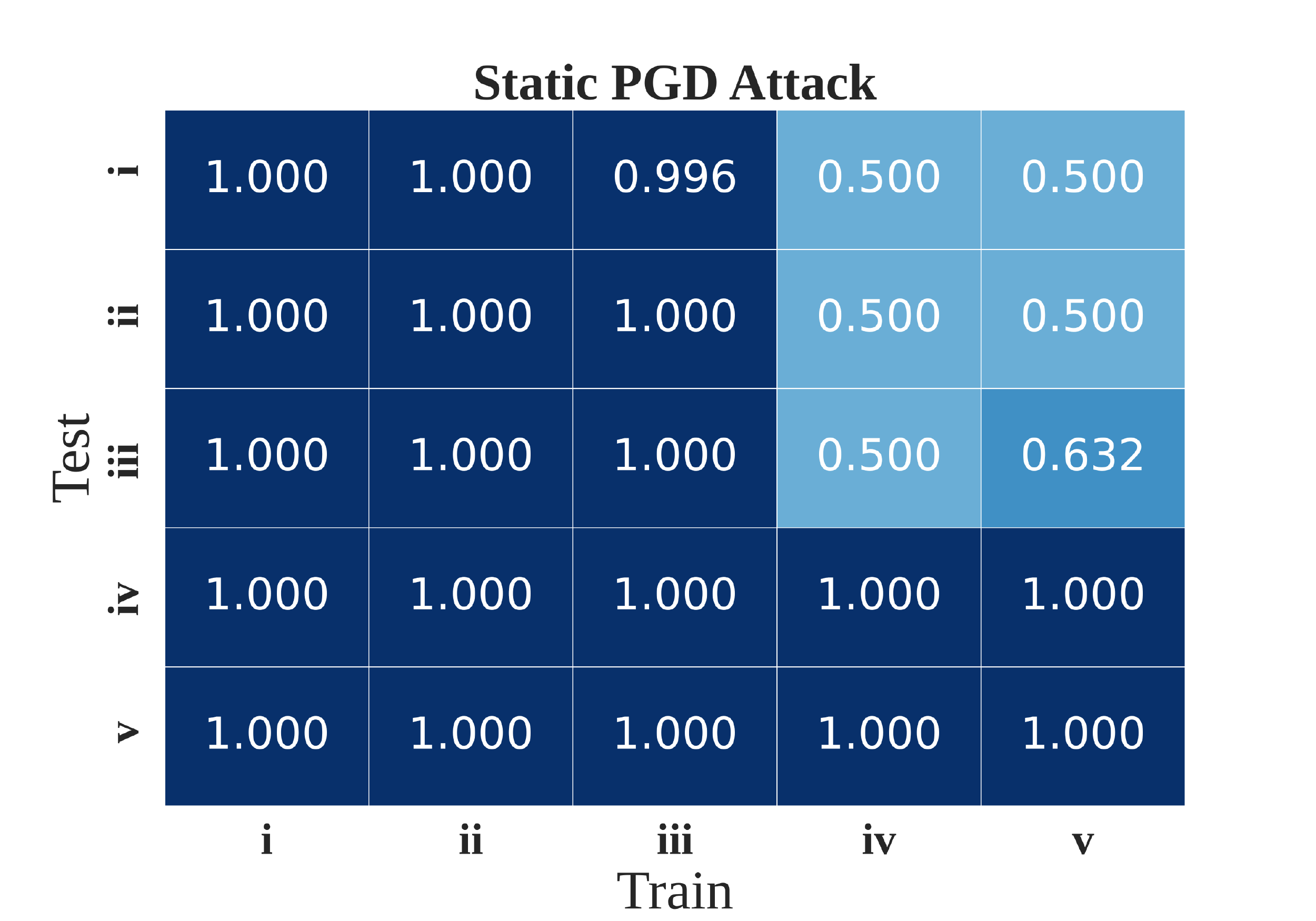} }}%
    \caption{AUC scores for static white-box PGD attacks on a detector-augmented ResNet18 network. The adversarial examples are generated from CIFAR-10, and the attacks vary in strength, as referenced in Table \ref{static-cifar10-attack-params}.}%
    \label{fig:static-resnet-pgd-cifar10}%
\end{figure}

In our ResNet18 implementation, we add a detector after the first block (layer 5). We find that our ResNet18 network appended with a detector is able to achieve robustness results on par with our augmented VGG19 network. Furthermore, we demonstrate that the ResNet18-based detection system improves state-of-the-art robustness when trained on a weaker PGD attack, which is discussed further in Section \ref{comparison-section}.  As shown in Fig. \ref{fig:static-resnet-pgd-cifar10}, the ResNet18 detector trained on weaker PGD attacks achieves an AUC score of 1.0 for each static attack. 

We additionally train a ResNet18 model on CIFAR-100 and add a detector after the first block (layer 5), finding that our detector is robust against adversarial examples generated from CIFAR-100. The robustness results in Table \ref{static-resnet-cifar100} demonstrate that our ANS-based detector is capable of detecting static PGD attacks with AUC scores of at least 0.999. Given the relative complexity of CIFAR-100, the results demonstrate that the ANS metric indeed provides a more structured and robust approach for adversarial detection, regardless of the number of model classes.

 \begin{table}[hbt!]
  \caption{AUC scores for static white-box PGD attacks (represented by \textit{n}, \textit{a}, \textit{e} parameters) on a detector-augmented ResNet18 network. The detector is trained on adversarial examples generated from CIFAR-100 with a relatively weak $L_{\infty}$ PGD attack with $n=7$, $\alpha=0.007$, and  $\varepsilon=0.125$.}
  \label{static-cifar100-table}
  \centering
  \resizebox{0.25\textwidth}{!}{\begin{tabular}{llll}%
    \cmidrule(r){1-4}
         {$n$}    & $\alpha$ & $\varepsilon$ & AUC      \\
    \midrule
    7 & 0.007 & 0.125 & 0.999   \\
    20 & 0.007 & 0.125 & 1   \\
    40 & 0.5 & 8.0 & 1   \\
    200 & 0.5 & 8.0 & 1   \\
    \bottomrule
    \label{static-resnet-cifar100}%
  \end{tabular}}
\end{table}

\begin{figure}[!htb]%
    \centering
    {{\includegraphics[width=6cm]{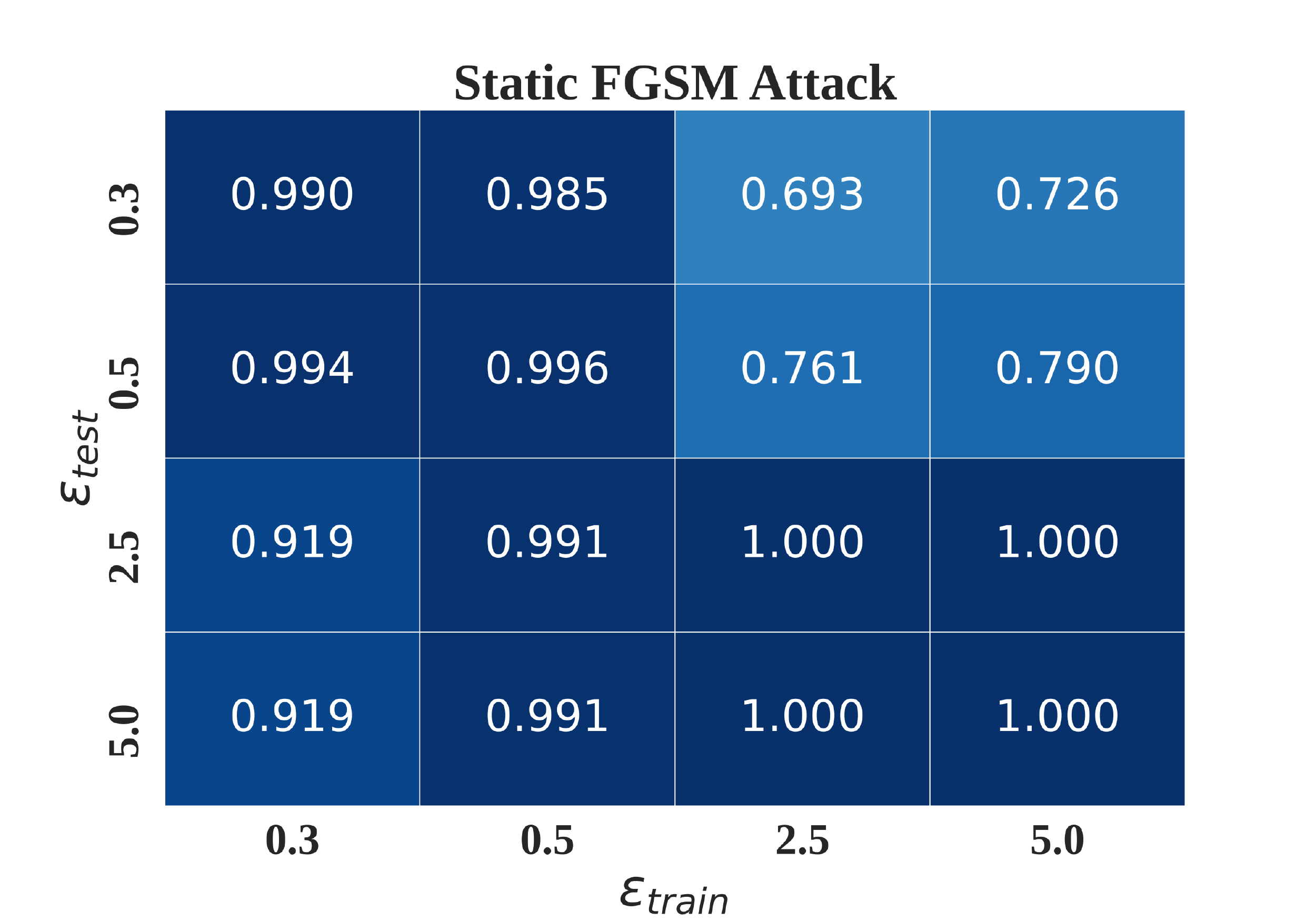} }}%
    \qquad
    {{\includegraphics[width=6cm]{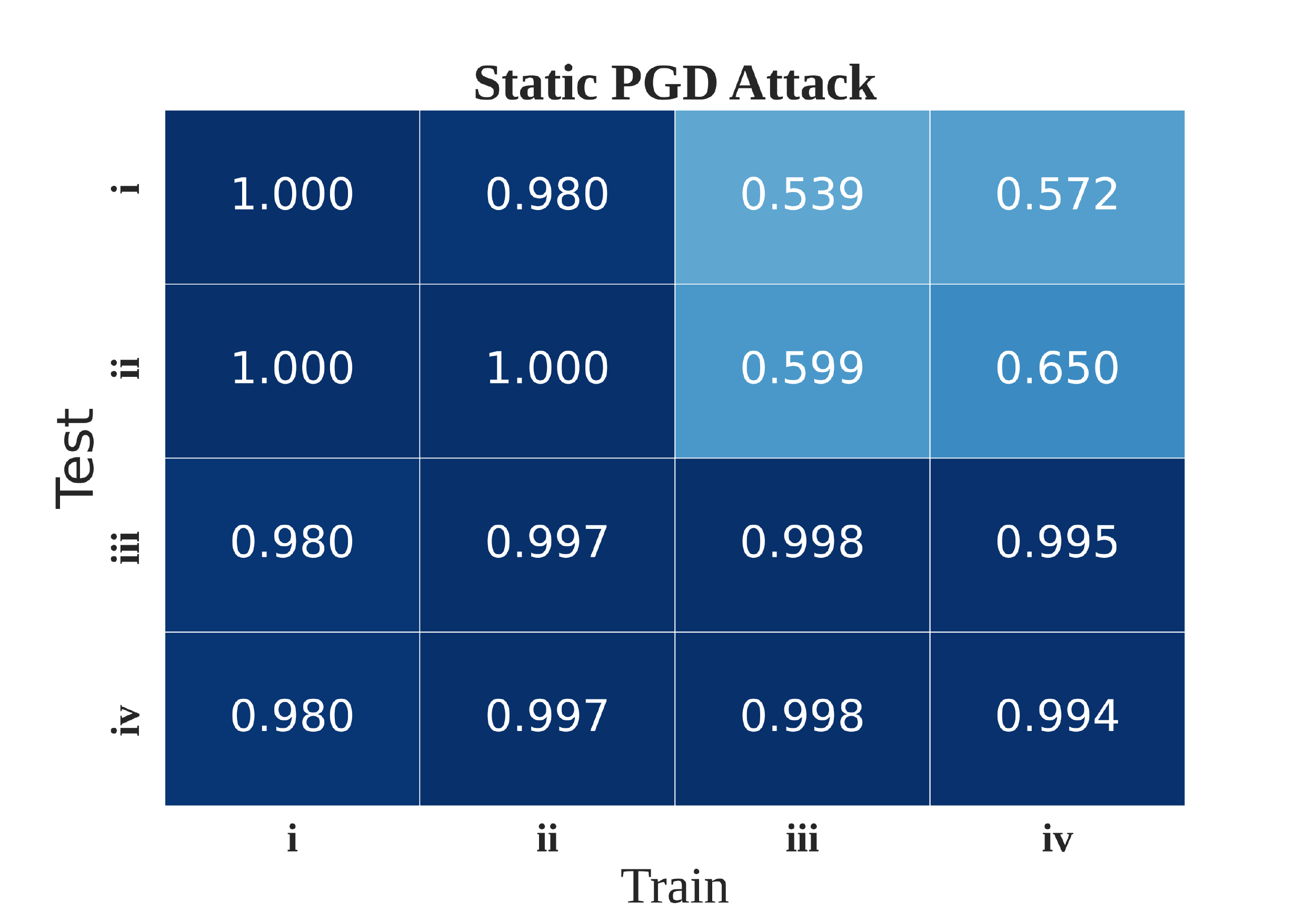} }}%
    \caption{AUC scores for static white-box FGSM and PGD attacks on a detector-augmented LeNet network. The adversarial examples are generated from MNIST, and the attacks vary in strength. The FGSM epsilon strengths are directly labeled in the figure, and the PGD attack labels correspond to the following parameters: \textbf{i.} $n=150$, $\alpha=0.007$, $\varepsilon=0.125$; \textbf{ii.} $n=300$, $\alpha=0.007$, $\varepsilon=0.125$; \textbf{iii.} $n=450$, $\alpha=0.06$, $\varepsilon=5.0$; \textbf{iv.} $n=600$, $\alpha=0.06$, $\varepsilon=5.0.$}%
    \label{fig:mnist-static}%
\end{figure}

Using the MNIST dataset, we trained a binary classifier on different attack strengths of FGSM and PGD to compare the robustness of detectors trained on each respective attack. The LeNet implementation we use consists of two convolutonal layers followed by two fully connected layers. Here, layer 2 is the high ANS layer, as inferred from a layer-wise ANS graph similar to those in Fig. \ref{fig:ans-vgg19}(a),(b). Fig. \ref{fig:mnist-static} includes AUC scores for detectors trained and tested at different FGSM and PGD attack strengths. For the FGSM attack, we train and test on attacks with varying epsilon strengths ($\varepsilon$), and for the PGD attacks, we train and test on the following parameters: the number of attack steps (\textit{n}), step size ($\alpha$), and attack strength ($\varepsilon$). The results demonstrate that the LeNet detector is able to detect adversaries generated from both FGSM and PGD attacks. Comparing the detectors trained on different attack types, the FGSM-trained detector performs better across the board, but the PGD-trained detector is consistently more robust when trained on weaker attacks.

For each dataset and model, we observe a similar trend where the detector is capable of generalizing to stronger attacks. This occurs due to a one-way transferability: detectors trained on stronger attacks are more likely to classify adversarial activations from weaker attacks as clean because the training attack strength sets a threshold for adversarial activations, whereas detectors trained on weaker attacks are able to recognize highly-perturbed activations as adversarial because these activations are even more distinct from the clean activations than the training adversarial activations are. Furthermore, we find that training the detector on weaker attacks gives rise to a detection system that is robust against a wider range of attack strengths.

\begin{table}[!htb]
    \caption{Training and testing hyperparameters for PGD dynamic white-box attacks on CIFAR-10, which vary in the number of attack steps (\textit{n}), step size (\textit{a}), and attack strength (\textit{e}), as shown below.}
    \begin{minipage}{.5\linewidth}
      VGG19
      \centering
  {\begin{tabular}{*{5}{p{0.6cm}}}%
    \cmidrule(r){1-4}
         Attack    & Step Size ($n$) & Step Width ($\alpha$) & Epsilon ($\varepsilon$)      \\
    \midrule
    i &  7 & 0.007 & 0.125    \\
    ii &  20 & 0.007 & 0.125     \\
    iii & 100 & 0.007 & 0.125     \\
    iv &  40 & 0.5 & 8.0    \\
    v &  500 & 0.5 & 8.0   \\
    \bottomrule
  \end{tabular}}
    \end{minipage}%
    \begin{minipage}{.5\linewidth}
      \centering
        ResNet18
    {\begin{tabular}{*{5}{p{0.6cm}}}%
    \cmidrule(r){1-4}
         Attack    & Step Size ($n$) & Step Width ($\alpha$) & Epsilon ($\varepsilon$)      \\
    \midrule
    i &  7 & 0.007 & 0.125    \\
    ii &  20 & 0.007 & 0.125     \\
    iii & 100 & 0.007 & 0.125     \\
    iv &  40 & 0.5 & 8.0    \\
    v &  200 & 0.5 & 8.0   \\
    \bottomrule
  \end{tabular}}
    \end{minipage}
\label{dynamic-cifar10-attack-params}%
\end{table}

\subsection{Dynamic White-Box Attack}
In a dynamic white-box attack setting, the adversary has knowledge of the CNN, as well as the detector, and is capable of perturbing both the images and the intermediate activations. This threat model assumes that the adversary attacks the input images, then additionally attacks the corresponding intermediate activations prior to the activations propagating through the detector. In the following simulations, we consider the attack all-or-nothing, meaning we either attack both the input images and activations or neither.  It is important to note that the dynamic attack is much more difficult to implement practically; unlike the static white-box adversary, which only provides an adversarial input to the network, the dynamic white-box adversary additionally needs to access and modify the neural network's intermediate values. In the following dynamic white-box experiments, we evaluate detector robustness against CIFAR-10 for VGG19 and ResNet18 models with the attack hyperparameters detailed in Table \ref{dynamic-cifar10-attack-params}.

\begin{figure}[!htb]%
    \centering{{\includegraphics[width=6.5cm]{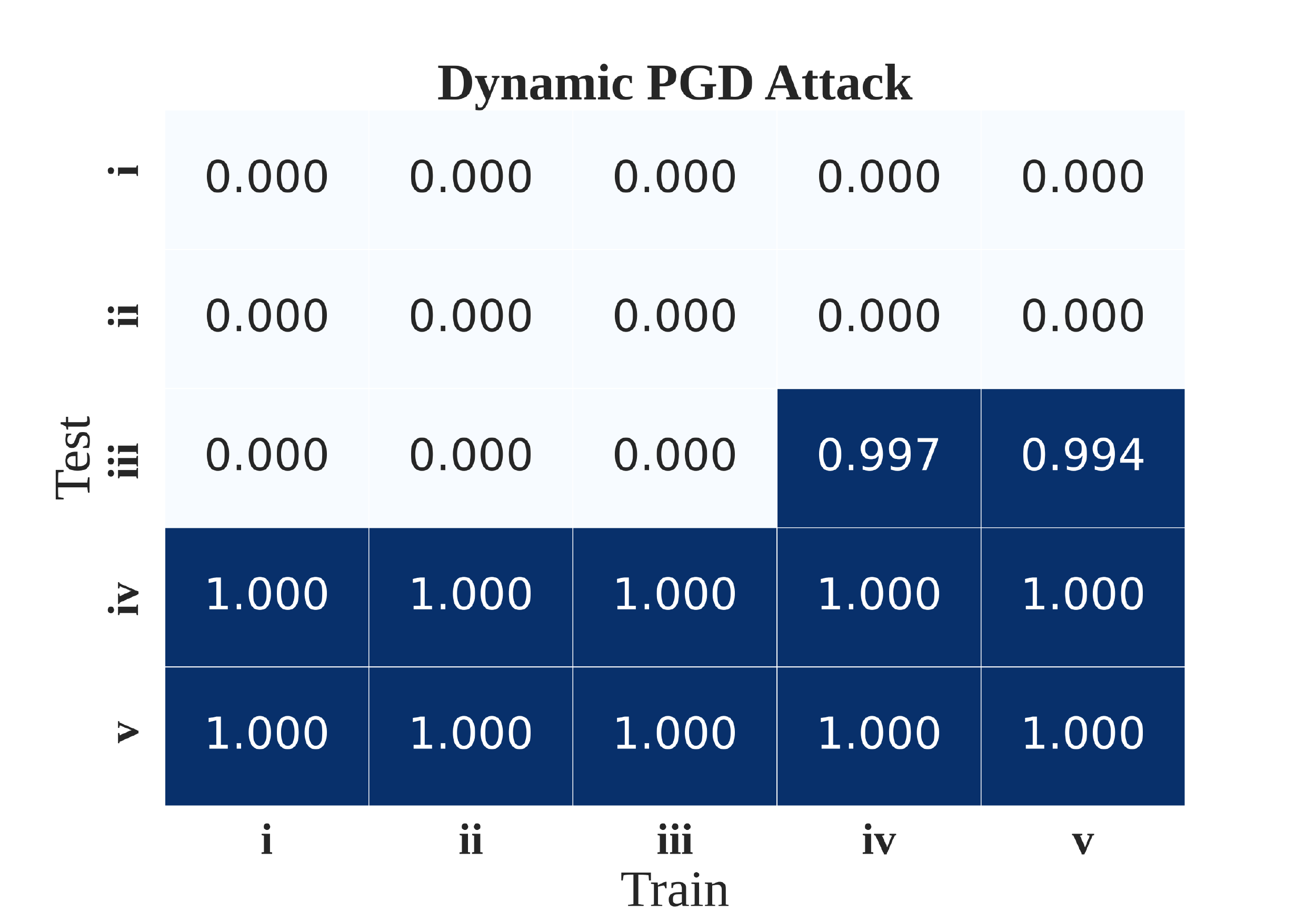} }}%
    \caption{AUC scores for dynamic white-box PGD attacks on a detector-augmented VGG19 network. The adversarial examples are generated from CIFAR-10, and the attacks vary in strength, as referenced in Table \ref{dynamic-cifar10-attack-params}.}%
    \label{fig:dynamic-vgg-pgd-cifar10}%
\end{figure}

\begin{figure}[!htb]%
    \centering{{\includegraphics[width=6.5cm]{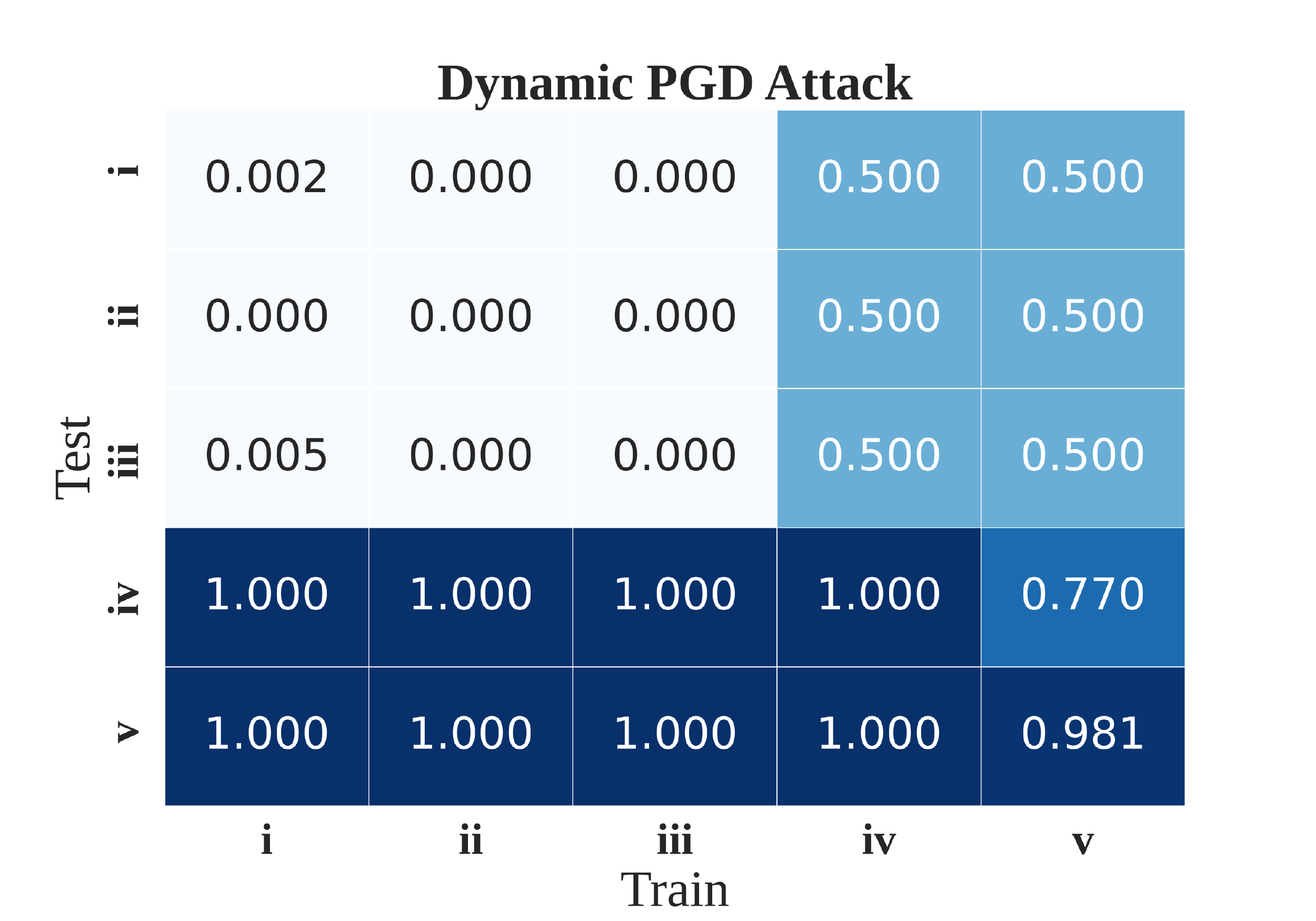} }}%
    \caption{AUC scores for dynamic white-box PGD attacks on a detector-augmented ResNet18 network. The adversarial examples are generated from CIFAR-10, and the attacks vary in strength, as referenced in Table \ref{dynamic-cifar10-attack-params}.}%
    \label{fig:dynamic-resnet-pgd-cifar10}%
\end{figure}

Fig. \ref{fig:dynamic-vgg-pgd-cifar10} and Fig. \ref{fig:dynamic-resnet-pgd-cifar10} display detector robustness against VGG19 and ResNet18 augmented networks, respectively. For both networks, we find that the detector is vulnerable to dynamic attacks, especially when the activations are modified with weaker PGD attacks. We observe AUC scores of 0 in the dynamic PGD attack cases where the attacks are able to completely fool the detector; likewise, these detectors generally predict the same output class for each image, resulting in accuracy scores close to 0.5. Interestingly, we also find that the detectors are able to achieve perfect or near-perfect robustness for stronger PGD attacks, even in the dynamic attack case. This is likely because the adversarial activations are drastically different from the clean activations to the extent that a PGD activation attack of such strength cannot fool the detector.


\subsection{Dynamic White-Box Attack with Adversarial Training of Detector}
In the following attack simulations, we explicitly train the detector with adversarial activations, unlike the previous dynamic white-box attack scenario, where the detectors are not trained with adversaries. Furthermore, we demonstrate that the detectors are robust against dynamic white-box attacks after they are trained on adversarial intermediate activations. We use an approach similar to that of Metzen et al. \cite{metzen2017detecting} to train the detector on dynamic white-box adversaries: as we train the detector, we generate PGD attacks on the detector layer's adversarial activations with 0.5 probability. In the following experiments, we demonstrate that we can combat dynamic attacks with high detectability by training the detector on both clean and adversarial activations.

\begin{figure}[!htb]%
    \centering{{\includegraphics[width=6.5cm]{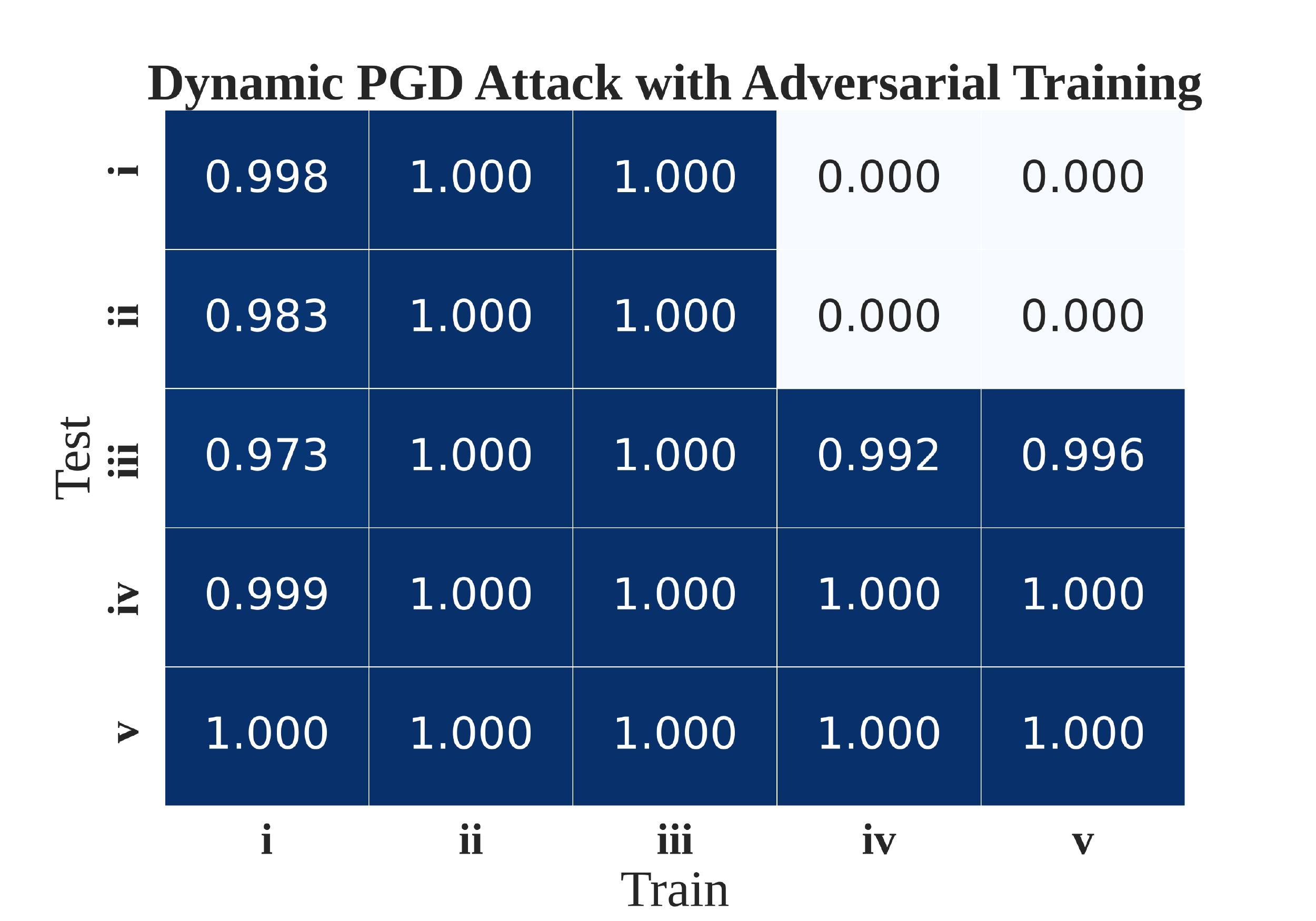} }}%
    \caption{AUC scores for dynamic white-box PGD attacks on a detector-augmented VGG19 network, with adversarial training of the detector. The adversarial examples are generated from CIFAR-10, and the attacks vary in strength, as referenced in Table \ref{dynamic-cifar10-attack-params}.}%
    \label{fig:dynamic-training-vgg-pgd}%
\end{figure}

\begin{figure}[!htb]%
    \centering{{\includegraphics[width=6.5cm]{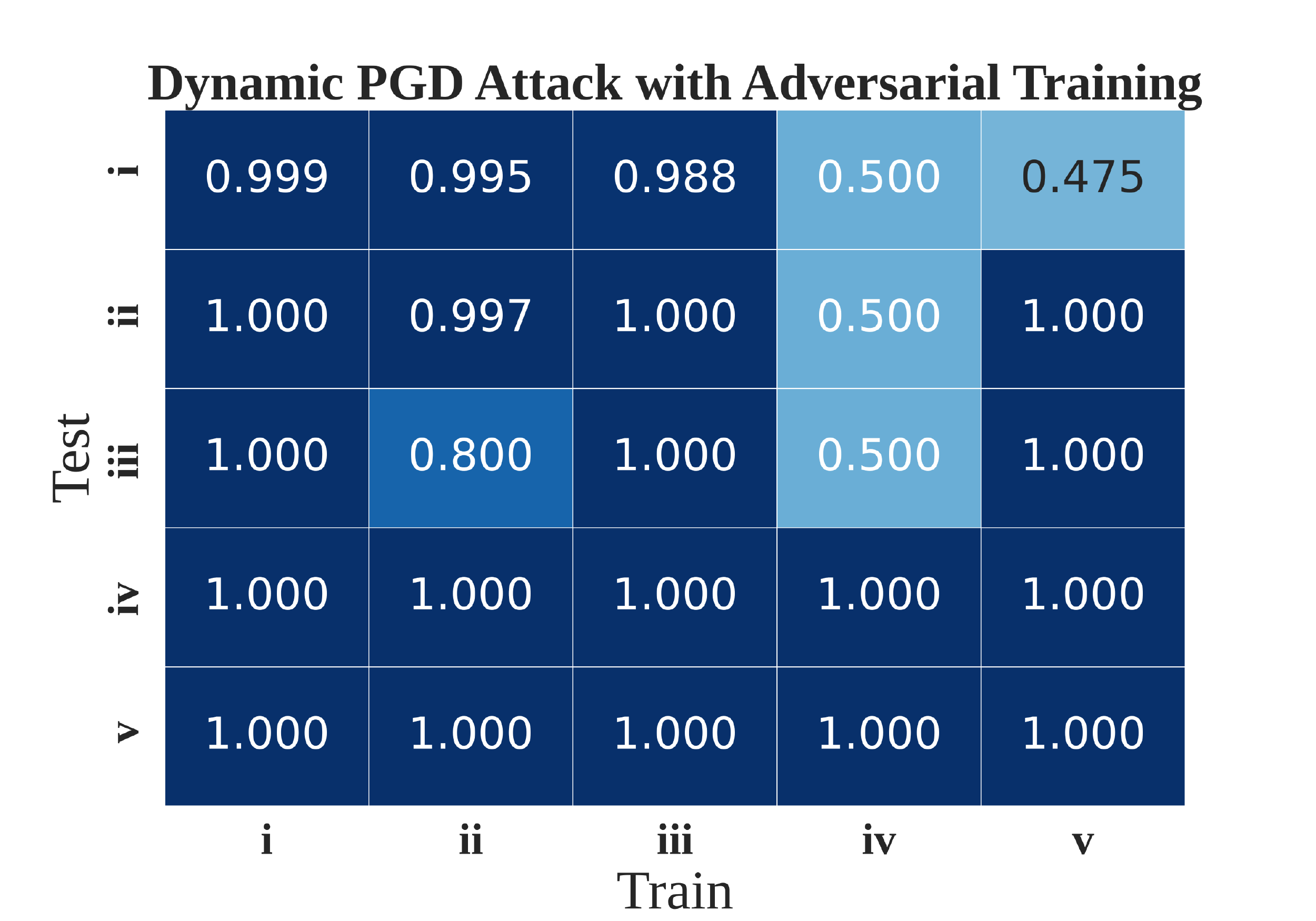} }}%
    \caption{AUC scores for dynamic white-box PGD attacks on a detector-augmented ResNet18 network, with adversarial training of the detector. The adversarial examples are generated from CIFAR-10, and the attacks vary in strength, as referenced in Table \ref{dynamic-cifar10-attack-params}.}%
    \label{fig:dynamic-training-resnet-pgd}%
\end{figure}

As shown in Fig. \ref{fig:dynamic-training-vgg-pgd} and Fig. \ref{fig:dynamic-training-resnet-pgd}, both the VGG19 and ResNet18 networks are highly capable of identifying dynamic white-box attacks after the detectors are trained on adversarial activations. Additionally, we simulate dynamic white-box attacks with CIFAR-100 and find that the detector with adversarial training is robust against dynamic white-box attacks. Table \ref{dynamic-training-resnet-cifar100} includes the CIFAR-100 experiment results, demonstrating that the detector achieves AUC scores of at least 0.999 against dynamic PGD attacks. Across each dataset and model, the results indicate that the detectors are more robust against dynamic white-box attacks when trained on weaker PGD attacks; these findings are consistent with the attack strength generalizability trend discussed in Section \ref{static-whitebox-section}. Based on our observations, we encourage future research to include transferability results for a range of strong and weak attacks and to consider training on weaker attacks.

 \begin{table}[hbt!]
  \caption{AUC scores for dynamic white-box PGD attacks (represented by \textit{n}, \textit{a}, \textit{e} parameters) on a detector-augmented ResNet18 network, with adversarial training of the detector. The detector is trained on adversarial examples generated from CIFAR-100 with a relatively weak $L_{\infty}$ PGD attack with $n=7$, $\alpha=0.007$, and  $\varepsilon=0.125$.}
  \label{cifar100-table}
  \centering
  \resizebox{0.25\textwidth}{!}{\begin{tabular}{llll}%
    \cmidrule(r){1-4}
         {$n$}    & $\alpha$ & $\varepsilon$ & AUC      \\
    \midrule
    7 & 0.007 & 0.125 & 0.999   \\
    20 & 0.007 & 0.125 & 1   \\
    40 & 0.5 & 8.0 & 1   \\
    200 & 0.5 & 8.0 & 1   \\
    \bottomrule
    \label{dynamic-training-resnet-cifar100}%
  \end{tabular}}
\end{table}

\subsection{Black-Box Attacks}
We also simulate black-box attacks on the CIFAR-10 and CIFAR-100 datasets. Similar to the method proposed by Papernot et al. \cite{papernot2016ablackbox} for generating black-box adversarial examples, we train a substitute model to craft adversarial examples that are used to attack the target model. Specifically, we use VGG19 substitute models to generate adversarial examples, then attack detector-augmented ResNet18 networks. In our black-box threat model, we assume the adversary has access to the full testing data, however in a real-world scenario it is more likely that an attacker would have to create their own synthetic dataset by observing the output labels for a contrived set of input images. 

Using the CIFAR-10 dataset, we first evaluate model robustness for black-box attack scenarios without any defense mechanisms to establish a baseline of the attack strength. Averaging across five trials without a detector, we find that
adversarial examples generated from a weak PGD black-box attack ($n=7$, $\alpha=0.007$, $\varepsilon=0.125$) using a VGG19 model degrades the ResNet18 model accuracy to $40.42\%$; on the contrary, a white-box PGD attack is much more effective, reducing the test accuracy to $0.21\%$. Additionally we find that a stronger PGD black-box attack ($n=200$, $\alpha=0.5$, $\varepsilon=8$) degrades the model to $10.00\%$, whereas a stronger white-box attack fools the network entirely, resulting in a test accuracy of $0\%$. In the case of stronger PGD black-box attacks, we find the minimum attack accuracy to be $10.00\%$ because the ResNet18 model predicts each adversarial example to be the same class. 

However, once we append a detector to the ResNet18 model, the network is capable of identifying adversarial examples generated from a black-box attack. As shown in Table \ref{blackbox-cifar10}, in each of the four black-box attack scenarios tested, the detector is able to perfectly distinguish between adversarial and clean examples, as reflected by an AUC score of 1; accordingly, the detector also classifies adversarial examples with at least $99.7\%$ accuracy. Given the difficulty of identifying black-box adversarial examples, these results further demonstrate the robustness of an ANS-based detector.

\begin{table}[hbt!]
  \caption{Accuracy and AUC scores for black-box PGD attacks (represented by \textit{n}, \textit{a}, \textit{e} parameters) on a detector-augmented ResNet18 network. The detector is trained on adversarial examples generated from CIFAR-10 with a relatively weak $L_{\infty}$ PGD attack with $n=7$, $\alpha=0.007$, and  $\varepsilon=0.125$}
  \centering
  \resizebox{0.3\textwidth}{!}{\begin{tabular}{lllll}%
    \cmidrule(r){1-5}
         \textit{n} & \textit{a} & \textit{e}   & AUC & Accuracy     \\
    \midrule
    7 & 0.007 & 0.125 & 1 & 0.997  \\
   20 & 0.007 & 0.125 & 1 & 0.998  \\
   40 & 0.5 & 8.0 & 1 & 0.998  \\
   200 & 0.5 & 8.0 & 1 & 0.998  \\
    \bottomrule
    \label{blackbox-cifar10}%
  \end{tabular}}
\end{table}

 Additionally, we use the CIFAR-100 dataset to test detector robustness against black-box attacks for a more complicated task, and find that the detector-appended ResNet18 network is unable to detect black-box adversarial examples. As shown in Table \ref{blackbox-cifar100}, the detector performs worse than random in terms of AUC score and accuracy for each adversarial attack scenario tested. This can be explained by the fact that each attack had a false negative rate of about $25\%$ and a false positive rate equal to or close to $100\%$. Likewise, the results indicate that the detector is unable to distinguish between clean and adversarial examples on the CIFAR-100 dataset. Given the relatively small amount of training data in the CIFAR-100 dataset, it would be worthwhile for future research to investigate the detectability of adversarial examples on other large scale datasets, such as ImageNet.
 
 \begin{table}[hbt!]
  \caption{Accuracy and AUC scores for black-box PGD attacks (represented by \textit{n}, \textit{a}, \textit{e} parameters) on a detector-augmented ResNet18 network. The detector is trained on adversarial examples generated from CIFAR-100 with a relatively weak $L_{\infty}$ PGD attack with $n=7$, $\alpha=0.007$, and  $\varepsilon=0.125$}
  \centering
  \resizebox{0.3\textwidth}{!}{\begin{tabular}{lllll}%
    \cmidrule(r){1-5}
         \textit{n} & \textit{a} & \textit{e}   & AUC & Accuracy     \\
    \midrule
    7 & 0.007 & 0.125 & 0.364 & 0.430  \\
   20 & 0.007 & 0.125 & 0.252 & 0.402  \\
   40 & 0.5 & 8.0 & 0 & 0.377  \\
   200 & 0.5 & 8.0 & 0 & 0.377  \\
    \bottomrule
    \label{blackbox-cifar100}%
  \end{tabular}}
\end{table}

\subsection{Attack Generalizability}
We find that within an attack algorithm, detector robustness is typically generalizable to stronger attacks only. However, this does not imply that ANS-based detectors are unable to detect adversarial examples generated by other attack algorithms, even if the attack algorithm is less powerful than that which it is trained on. For example, although FGSM is considerably weaker than PGD, the results in Table \ref{resnet-to-fgsm-generalize} show that our PGD-trained detector is able to detect FGSM attacks with high discernability, achieving an AUC score of 1.0 and accuracy of 0.998 for each attack tested. 
 
 \begin{table}[hbt!]
  \caption{Generalizability across different static attack algorithms: robustness of a detector trained on PGD attacks (\textit{n} = 7, \textit{a} = 0.007, \textit{e} = 0.125) and attacked with varying epsilon strengths (\textit{e}) of FGSM.}
  \centering
  \resizebox{0.23\textwidth}{!}{\begin{tabular}{lllll}%
    \cmidrule(r){1-3}    
         \textit{e}   & AUC & Accuracy     \\
    \midrule
    0.3 & 1 & 0.998  \\
  0.5 & 1 & 0.998  \\
  2.5 & 1 & 0.998  \\
  5.0 & 1 & 0.998  \\
    \bottomrule
    \label{resnet-to-fgsm-generalize}%
  \end{tabular}}
\end{table}
 
 Likewise, the FGSM-trained detector also achieves high AUC and accuracy scores for PGD attacks, as shown in Table \ref{fgsm-to-resnet-generalize}. The discrepancy between AUC and accuracy for the weakest attack strength (\textit{n} = 7, \textit{a} = 0.007, \textit{e} = 0.125) is explained by the fact that the FGSM-trained detector is able to distinguish between adversarial and clean images. But, it does not classify attacked images as adversarial because the threshold imposes that the positive class predictions (adversarial examples) corresponds to a probability score greater than 0.5.  Follow up analyses reveal that for nearly each test example, the detector classifies the activations as clean because the positive class prediction scores are extremely low. However, there is a stark difference between the positive class probability scores for both the true negative and false negative cases: in the true negative case, about 99\% of the positive class probability scores fall in the range ($1\text{e-}6$, $1\text{e-}3$], and for the false negative case, over 99\% of the positive class probability scores fall in the range ($1\text{e-}3$, 1]. Thus, if a detector were to use a positive class prediction threshold of 0.001 instead of 0.5 for this attack scenario to distinguish between clean and adversarial examples, it would correctly detect adversarial examples with over 99\% accuracy. Furthermore, we demonstrate that accuracy follows AUC for stronger attack strengths, which supports our findings that robustness is generalizable between attack algorithms.

\begin{table}[hbt!]
  \caption{Generalizability across different static attack algorithms: robustness of a detector trained on FGSM attacks (\textit{e} = 3) and attacked with varying strengths of PGD, as represented by the \textit{n}, \textit{a}, \textit{e} parameters.}
  \centering
  \resizebox{0.3\textwidth}{!}{\begin{tabular}{lllll}%
    \cmidrule(r){1-5}
         \textit{n} & \textit{a} & \textit{e}   & AUC & Accuracy     \\
    \midrule
    7 & 0.007 & 0.125 & 0.999 & 0.504  \\
   20 & 0.007 & 0.125 & 1 & 1  \\
   40 & 0.5 & 8.0 & 1 & 0.981  \\
   200 & 0.5 & 8.0 & 1 & 1  \\
    \bottomrule
    \label{fgsm-to-resnet-generalize}%
  \end{tabular}}
\end{table}

\subsection{Comparison with State-of-the-Art Methods}
\label{comparison-section}
In addition to providing a more energy-efficient and structured approach to detecting adversarial examples, our ANS-based detector also improves previous methods in terms of robustness. In Table \ref{yin-comparison}, we compare our approach with the state-of-the-art detection method proposed by Yin et al. \cite{yin2020gat} and demonstrate that our ResNet18 detector improves detection on CIFAR-10. Likewise, in Table \ref{metzen-comparison}, we compare our method with Metzen et al. \cite{metzen2017detecting}, the approach that is most similar to ours, and show that our detector improves state-of-the-art detectors based on intermediate activations for CIFAR-10. In both comparisons, our ResNet18 detector is trained on a relatively weak PGD attack with $n=7$, $\alpha=0.007$, and $\varepsilon=0.125$.

\begin{table}[!hbt]
  \caption{AUC score comparison with Yin et al. \cite{yin2020gat} state-of-the-art detector for $L_{\infty}$ PGD white-box attack.}
  \centering
  \begin{tabular}{lllll}
    \\
    \cmidrule(r){1-3}
        $n$, $\alpha$, $\varepsilon$ & 40, $0.5$, 8  & 200, $0.5$, 8      \\
    \midrule
    State-of-the-art (Yin et al., 2020) & 0.955 & 0.950   \\
    Ours (ResNet18 detector) &  1 & 1     \\
    \bottomrule
    \label{yin-comparison}%
  \end{tabular}
\end{table}

\begin{table}[!hbt]
  \caption{Accuracy comparison with Metzen et al. \cite{metzen2017detecting}  state-of-the-art intermediate activations-based detector for iterative white-box attack.}
  \centering
  \begin{tabular}{lllll}
                    \\
    \cmidrule(r){1-3}   { $n$, $\alpha$, $\varepsilon$} & 10, 1, 2  & 10, 1, 4      \\
    \midrule
    State-of-the-art (Metzen et al., 2017) & 0.950 & 0.960   \\
    Ours (ResNet18 detector) &  0.998 & 0.998    \\
    \bottomrule
    \label{metzen-comparison}%
  \end{tabular}
\end{table}

To the best of our knowledge, there is no standard adversarial robustness benchmark on CIFAR-100, however we find that our CIFAR-100 detector performs better than prior detection-based methods do on less complex image classification datasets, including CIFAR-10 (Table \ref{yin-comparison}, Table \ref{metzen-comparison}). These results on CIFAR-100 validate that our ANS-based detector is robust against white-box attacks on more difficult classification tasks.

In addition to comparing robustness with state-of-the-art methods, we'd also like to note the energy advantages of our approach in relation to the large computational costs associated with previous detection-based designs. For detection-based approaches, the layer at which a network detects adversarial examples relative to the entire network depth impacts the computational overhead of the detection strategy. For example, Grosse et al. \cite{grosse2017detection} and Feinman et al. \cite{feinman2017detecting} perform classification at or after the final hidden layer, which is energy inefficient because it requires adversarial examples to be propagated through each DNN layer. While Metzen et al. [18] perform early adversarial detection by adding a detector to an intermediate layer, their methodology lacks a structured heuristic for choosing the detector location. Instead, the detector location is chosen after experimenting with various possible locations. This adds to the computational overhead and unpredictability. Likewise, Li and Li \cite{li2017adversarial} add cascade classifiers to each convolutional layer, Yin et al. \cite{yin2020gat} use detectors for each class, and Feinman et al. \cite{feinman2017detecting} use three distinct detectors. These works adds structural complexity to the network, incurring higher computational costs. Furthermore, the approach by Bhagoji et al. \cite{bhagoji2018enhancing} requires additional matrix multiplications to project inputs onto the principal components, which requires additional computations.

On the contrary, in the proposed ANS-based detection methodology, we aim to minimize the computational cost of detecting adversarial examples. We perform adversarial detection at shallow DNN layers via an early-exit strategy, which prevents unnecessary computations in deeper layers. Likewise, our ANS-based detection methodology provides a structured approach for selecting the most sensitive DNN layers. Based on ANS, we identify where to append the detector in the DNN, eliminating the need to experimentally determine the most sensitive layer. Additionally, we perform adversarial detection using a small neural network-based binary classifier to minimize the structural complexity of the detector. Lastly, this is the first work, to the best of our knowledge, that focuses on hardware-algorithm co-design to ensure that our detection system is pragmatically energy efficient. By implementing the detector end-to-end on hardware scalable CMOS accelerators, with quantization, we demonstrate that our methodology reduces the energy consumption of the entire system.

\subsection{Quantized Networks}
\label{quantized_nets}
Network quantization compresses the network by reducing the number of bits used to represent the network parameters, including weights and activations. By quantizing a neural network, we can reduce the energy consumption required to access the network weights and to compute and store the activations \cite{han2015compression}\cite{wu2016quantized}. Here, we explore how we can reduce the data access and computation energies without degrading network robustness against adversarial attacks in the proposed detector-based scheme. Consistent with the earlier experiments, we use a test dataset of 10,000 clean activations and 10,000 adversarial activations from CIFAR-10; the adversarial activations are generated from a PGD attack with $n=7$, $\alpha=0.007$, $\varepsilon=0.125$. 

\begin{figure}[!htb]%
    \centering{{\includegraphics[width=8cm]{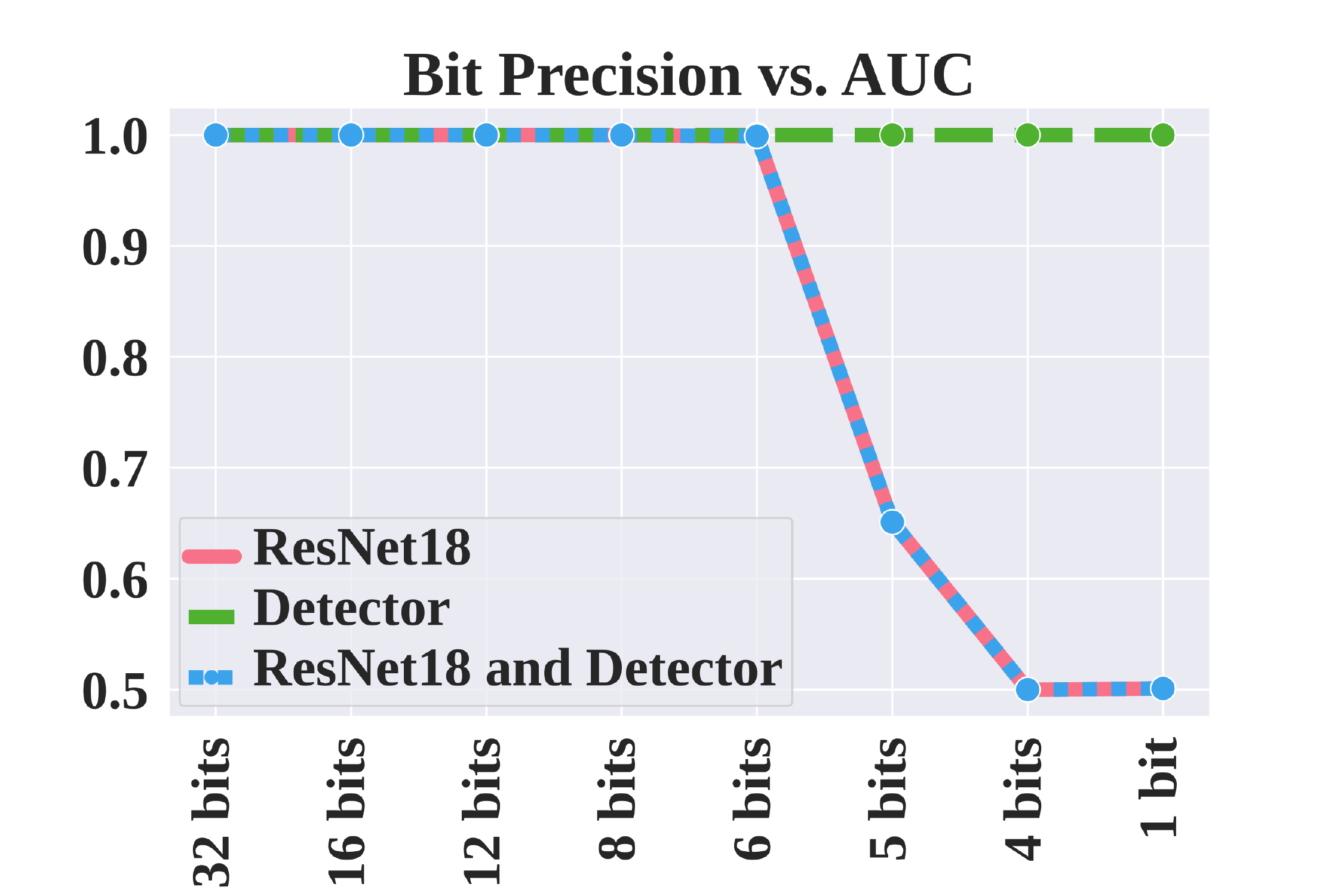} }}%
    \caption{AUC scores for various bit precisions after quantization of weights and activations for (a) ResNet18 only (b) the detector only and (c) both ResNet18 and the detector.}%
    \label{fig:network-auc-by-bit}%
\end{figure}

Fig. \ref{fig:network-auc-by-bit} reflects how the AUC score of the detector changes for varying bit precisions used to represent model weights and activations. We look at three different cases: 1) when only the ResNet18 network parameters are quantized, 2) when only the detector parameters are quantized, and 3) when both the ResNet18 and detector parameters are quantized. When we only quantize parameters for the ResNet18 network, we pass the activations from the quantized network to the detector and find that the AUC score of the detector begins to drop slightly (by 0.001) at a 6-bit precision for ResNet18 and more drastically for 5-bit precisions and below. On the contrary, only quantizing the detector parameters has a negligble effect on the AUC score; the AUC score remains 1.0 for each bit precision shown. Likewise, the AUC degradation trend for quantizing both the ResNet18 and detector parameters is identical to that of the case where only the ResNet18 network is quantized. Due to the fact that the detector robustness is unaffected when the detector is quantized to 1-bit precision, as seen in the second case, the AUC degradation in the third case (when both the ResNet18 model and detector are quantized) can be attributed to the effect of quantization on the ResNet18 network alone. Based on these results, we conclude that we can reduce energy consumption and maintain a strong AUC score by using a 6-bit precision for the ResNet18 and 1-bit precision for the detector. However, before assigning the final bit-precisions for the network and detector, further analysis needs to be performed as explained below.

\begin{figure}[!htb]%
    \centering{{\includegraphics[width=8cm]{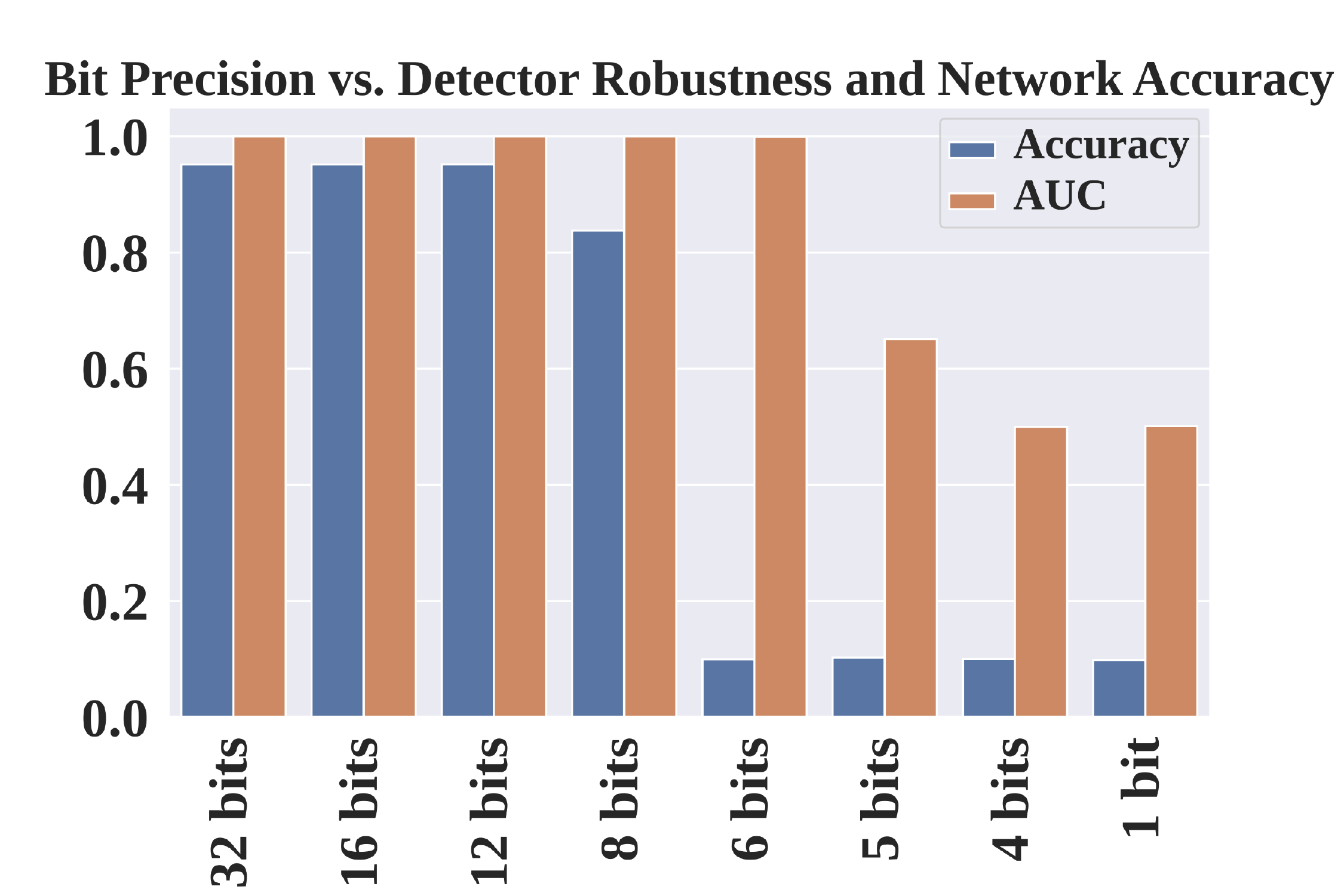} }}%
    \caption{Detector AUC scores compared to ResNet18 accuracy shown for various bit precisions.}%
    \label{fig:detector-auc-by-bit}%
\end{figure}

When evaluating robustness with quantized networks, we must also consider how reduced bit precisions affect the network accuracy for classifying images. Fig. \ref{fig:detector-auc-by-bit} compares changes in ResNet18 network accuracy and detector AUC scores for different bit precisions. We find that although the AUC score degrades for precisions below 6 bits when both the network and detector are quantized, the ResNet18 network accuracy declines for quantized networks smaller than 12 bits. Thus, we use a 12-bit precision for the ResNet18 model and 1-bit precision for the detector in order to maintain the high accuracy as well as robustness of ResNet18. 

\section{Hardware Implementation}
In this section, we give a detailed explanation of the hardware accelerator implemented for the proposed network-detector architecture described above and additionally demonstrate the energy and compute efficiency of our design under different testing scenarios. 
\subsection{Experimental Setup}
\begin{figure}[!htb]%
    \centering{{\includegraphics[width=6cm]{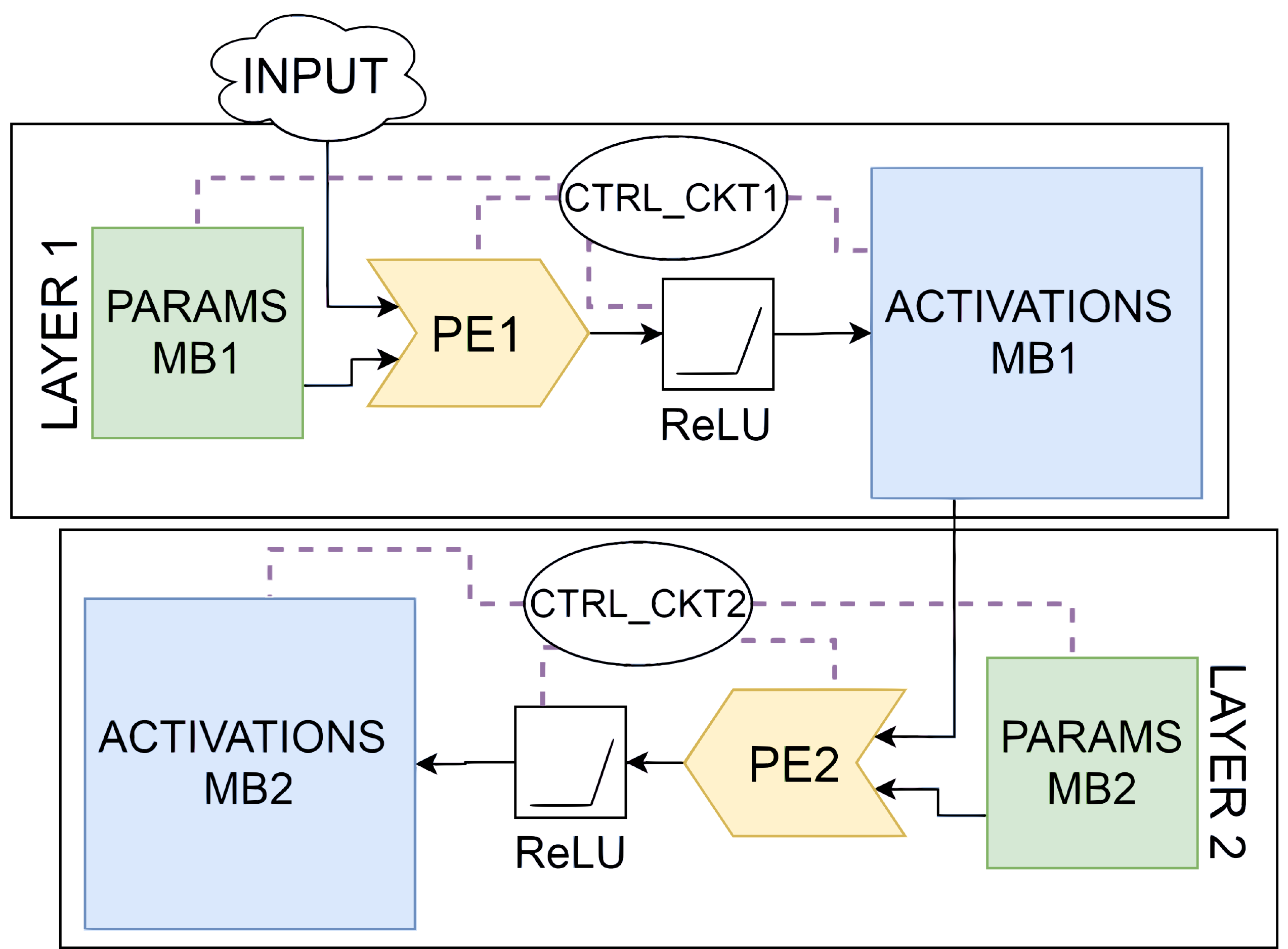} }}%
    \caption{Block diagram of the hardware accelerator architecture.}%
    \label{accelerator}%
\end{figure}

We design a MAC accelerator shown in Fig. \ref{accelerator} using system verilog descriptions capable of supporting variable data (parameters and activations) precision for each layer. The accelerator can support up to 16-bit data precision. To incorporate hardware scalability, we design the MAC unit to support DG and DVAFS computation paradigms which enable energy efficient approximate computation \cite{moonsdg} \cite{moonsdvafs}. For accurate energy estimations, the design was implemented using 45nm CMOS technology with conservative power models having a nominal supply voltage of 1V. The accelerator can be operated in the following different modes, where N $\in$ [1,16]. \begin{enumerate}
    \item \textbf{Basic Mode}: N-bit MAC operations are carried out without employing any hardware scalibility methods.
    \item \textbf{DG Mode}: N-bit MAC operations are computed using only MSBs according to the DG convention.
    \item \textbf{DVAFS Mode}: Similar to DG Mode, here MACs are carried out with sub-word level parallelism (of 2X 8-bit and 4X 4-bit levels) and frequency scaling for more energy optimized operations. 
\end{enumerate}
The layers in the accelerator are mapped one-to-one with the layers of the proposed model. Each layer has its own parameter memory banks and activation memory banks, along with processing elements (PE) and control circuit ($ctrl\_ckt$). Memory banking helps minimize the time and energy required during data transfer and facilitates parallel implementation. The PE contains the MAC unit that performs MAC operations on the input data. The $ctrl\_ckt$ serves as the principal component, dictating signals to ensure 1) proper switching between the three modes of operations and 2) accurate data-flow between the memory banks to the processing elements and storage of the activation values.

\subsection{Energy Analysis}
\label{energy_analysis}
Using the projection of a 45nm CMOS SPICE accelerator implementation, we calculate the energy consumed during a single MAC operation (in all the three modes mentioned previously) and memory access, as shown in Table \ref{energy_consumption}. Here, the energy values provided correspond to 32-bit data memory access and MAC operations. This helps to approximate the energy consumption for any other data precision $k_b$ as mentioned in Table \ref{energy_consumption}. We will be using these MAC energy values for all our energy analysis experiments. 
\begin{table}[!hbt]
  \caption{Energy analysis for single MAC and data-memory access.}
  \label{energy_consumption}
  \centering
  \begin{tabular}{lllll}
    \toprule
    \cmidrule(r){2-3}
         Operation & Notation & Energy (pJ)     \\
    \midrule
    $k_b$ bit Memory access & $E_{A,k_b}$ & 2.5$k_b$   \\
    32 bit MULT 32 bit &  $E_{M,I}$ & 3.1     \\
    32 bit ADD 32 bit &  $E_{Add,I}$ & 0.1     \\
    $k_b$ bit MAC INT &  $E_{C,k_b}$ & (3.1 * $k_b$)/ 32) + 0.1    \\
    \bottomrule
  \end{tabular}
\end{table}

\begin{figure}[!htb]%
    \centering{{\includegraphics[width=7.5cm]{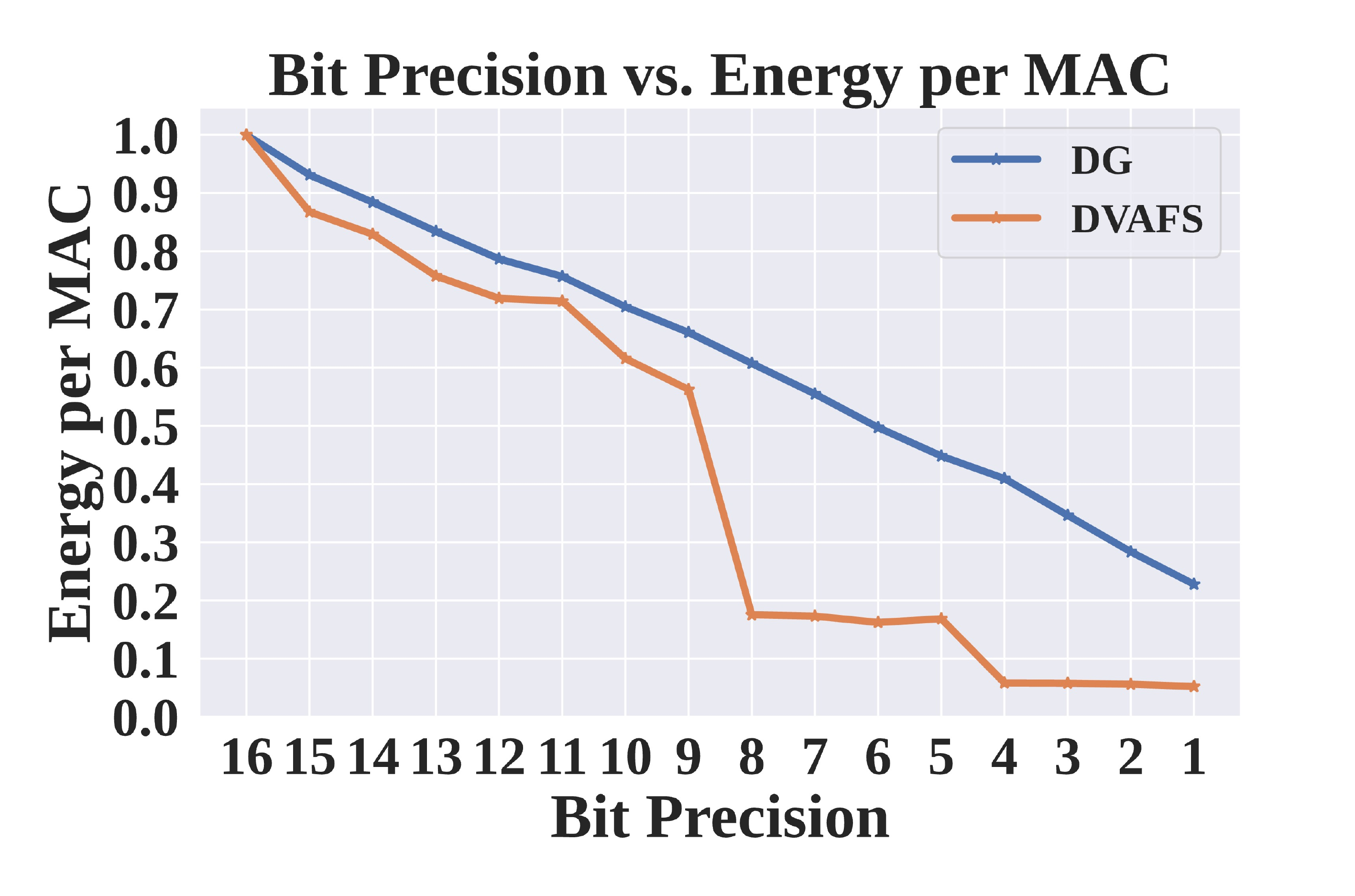} }}%
    \caption{Normalized energy per MAC against data precision for DG and DVAFS paradigms.}
    \label{dg_dvafs}%
\end{figure}

To evaluate MAC energy in DG and DVAFS paradigms, we vary the data precision in the range [1,16] and obtain the MAC energy per computation. The MAC energy values are then normalized with respect to MAC energy corresponding to the 16-bit data precision shown in Fig. \ref{dg_dvafs}. We use these normalized values to compare the energy efficiency of our proposed model with baseline models as explained later in this section. In the case of DVAFS for 8-bit and 4-bit precision, it is interesting to note the steep fall in the energy per MAC. This can be attributed to the fact that our accelerator implements $2\times8-bits$ and $4\times4-bits$ sub-word level parallelism for DVAFS. Hence, at 8-bit and 4-bit precision, the frequency (\textit{f}) can be further decreased to 0.5\textit{f} and 0.25\textit{f}, respectively, without affecting the net throughput; this also incurs a larger dip in the energy per MAC.

Typically, energy consumption can be directly related to two major aspects in a DNN accelerator: MAC operation energy and memory access energy. In a convolutional layer, the number of MAC operations is formulated as: $$N_{MAC} = N_I \times N_O \times M_O^{2} \times k^{2}$$ where $N_I$ is the number of input channels, $N_O$ is the number of output channels, $M_O$ is the output channel size, and $k$ is the kernel size. Likewise, the MAC energy for a layer having $k_b$-bit data precision can be calculated as: $$E_{MAC} = N_{MAC}\times E_{C,k_b}$$

To analyze the quantization energy reduction, we compare the MAC energies of a homogeneous 16-bit detector-appended network with a 12-bit quantized network with 1-bit quantized detector architecture. At iso-accuracy for clean inputs, the quantized detector-augmented VGG19 network architecture consumes 20\% lower MAC computation energy than the 16-bit architecture. Similarly, for ResNet18 architecture, the quantized network consumes 17\% lower MAC computation energy compared to the 16-bit network.  

However, the goal of our energy analysis experiments is to demonstrate the energy efficient adversarial example detection capability of our proposed detector-augmented network architecture. For this, we consider a \textit{baseline} model which is the standalone 12-bit quantized main network. The energy efficiency of adversarial detection is shown by subjecting the baseline and proposed detector model to three different testing scenarios: 1) No adversarial inputs 2) 99\% adversarial composition, and 3) 1\% adversarial composition for the CIFAR-10 dataset. It must be noted that the energy consumed by the \textit{baseline} model will be constant for all three cases because there is no early detection. Since our experiments in Section \ref{experiments-section} demonstrate that our detector-augmented network is capable of achieving an AUC score of 1 in static and dynamic attack scenarios, we assume that any adversarial example is detected and further propagation is terminated.  

Table \ref{E_MAC_resnet18} demonstrates the MAC energy efficiency of our approach for the detector-appended ResNet18 network architecture, where the binary detector is placed after the fifth layer (first block) of the ResNet18 main network. For the case when there are no adversarial examples, our proposed model consumes slightly more energy when compared to \textit{baseline} because of the detector network overhead. In the case when the concentration of adversarial examples in the test dataset is significant, our model is more energy efficient due to the early-exit strategy. Likewise, there is a greater reduction in energy consumption for larger concentrations of adversarial examples. Using a similar methodology, we perform MAC computation energy analysis on a detector-appended VGG19 network architecture. Here, the detector is placed at the end of the seventh layer. The results are shown in Table \ref{E_MAC_vgg19}.
\begin{table}[!hbt]
  \caption{MAC Energy results for 12-bit ResNet18 network with 1-bit detector compared with baseline model.}
  \label{E_MAC_resnet18}
  \centering
  \resizebox{0.45\textwidth}{!}{
  \begin{tabular}{ccccc}
    \toprule
    \cmidrule(r){2-3}
         & \begin{tabular}[c]{@{}c@{}}$E_{MAC,STD}$\\(J)\end{tabular} & \begin{tabular}[c]{@{}c@{}}$E_{MAC,DG}$\\(J)\end{tabular} & \begin{tabular}[c]{@{}c@{}}$E_{MAC,DVAFS}$\\(J)\end{tabular} & Reduction   \\
    \midrule
    12-bit Baseline & \num{6.93} & \num{4.32} & \num{3.94} & 1X\\
    No Adversaries & \num{6.97} & \num{4.37} & \num{3.96} & $1.01\times$ \\
    99\% Adversaries & \num{2.25} & \num{1.42} & \num{1.27} & $0.32\times$\\
    1\% Adversaries & \num{6.92} & \num{4.34} & \num{3.93} & $0.99\times$\\
    \bottomrule
  \end{tabular}
  }
\end{table}

\begin{table}[!hbt]
  \caption{MAC Energy results for 12-bit VGG19 network with 1-bit detector compared with baseline model.}
  \label{E_MAC_vgg19}
  \centering
  \resizebox{0.45\textwidth}{!}{
  \begin{tabular}{ccccc}
    \toprule
    \cmidrule(r){2-3}
         & \begin{tabular}[c]{@{}c@{}}$E_{MAC,STD}$\\(J)\end{tabular}  & \begin{tabular}[c]{@{}c@{}}$E_{MAC,DG}$\\(J)\end{tabular} & \begin{tabular}[c]{@{}c@{}}$E_{MAC,DVAFS}$\\(J)\end{tabular} & Reduction   \\
    \midrule
    12-bit Baseline & \num{5.02} & \num{3.13} & \num{2.86} & 1X\\
    No Adversaries & \num{5.04} & \num{3.14} & \num{2.87} & $1.01\times$\\
    99\% Adversaries & \num{2.44} & \num{1.53} & \num{1.39} & 0.49X\\
    1\% Adversaries & \num{5.01} & \num{3.13} & \num{2.84} & $0.99\times$\\
    \bottomrule
  \end{tabular}
  }
\end{table}

\begin{figure}[!htb]%
    \centering{{\includegraphics[width=7.5cm]{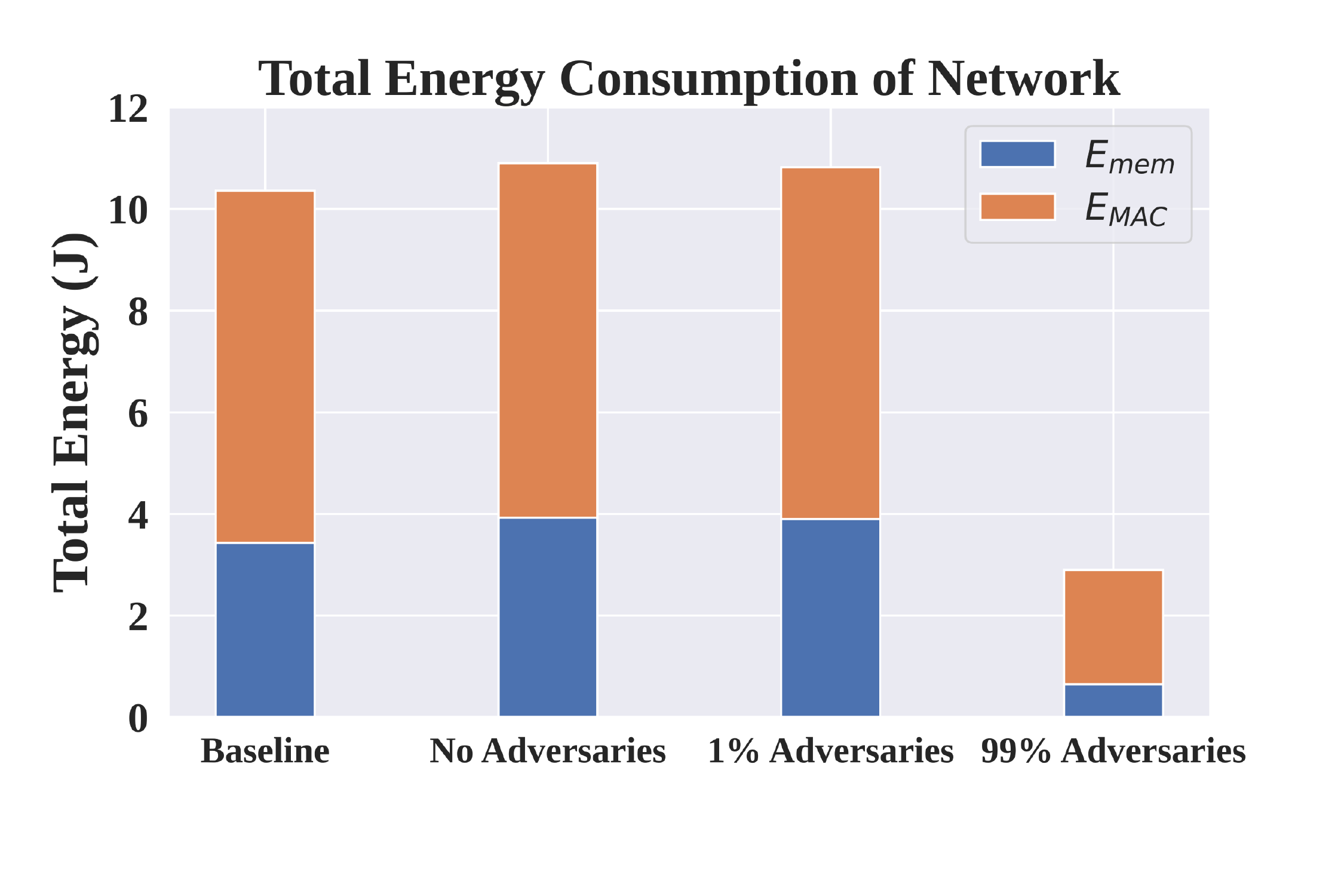} }}%
    \caption{Total energy consumed by the detector-appended ResNet18 architecture.}
    \label{tot_ener}%
\end{figure}

Another important energy contributor is the \textit{memory access energy}. This is the energy cost per access of the memory for read and write operations. The number of memory accesses for a CNN layer can be estimated using the following formula: $$N\_Access_{mem} = N_I \times M_I^{2} + N_I \times k^{2}$$
Using this, we can find the memory access energy $E_{mem}$, for a layer having $k_b$ bits data precision using,
$$E_{mem} = N\_Access_{mem} \times E_{A,k_b}$$
where $M_I$ is the input channel size, $E_{A,k_b}$ is the energy cost per $k_b$ bit memory access, and the other variables follow the previous notation from the MAC operations and energy formulas. Table \ref{E_mem} shows the total memory access energy for the 12-bit quantized VGG19 and ResNet18 networks with respective 1-bit quantized detectors attached at the end of the chosen layer. Our findings demonstrate that in the case when the concentration of adversarial examples in the test dataset is 1\%, the memory access energy is slightly higher than the baseline model, unlike the $E\_MAC$ case. This can be attributed to the fact that the detector involves about $~10\times$ higher number of memory accesses than the number of MAC computations. Finally, to obtain the total energy consumed by the 12-bit CNN with 1-bit detector quantized network, we combine the MAC computation energy and memory access energy values for both the detector-appended VGG19 and ResNet18 network architectures. For simplicity, we only show the energy plots for the detector-appended ResNet18 architecture in Fig. \ref{tot_ener}.
\begin{table}[!hbt]
  \caption{Memory access energy results for both ResNet18 and VGG19 networks with detector.}
  \label{E_mem}
  \centering
  \resizebox{0.45\textwidth}{!}{
  \begin{tabular}{ccccc}
    \toprule
    \cmidrule(r){2-3}
         & \begin{tabular}[c]{@{}c@{}}$E_{mem,ResNet18}$\\(J)\end{tabular}  & \begin{tabular}[c]{@{}c@{}}$E_{mem,VGG19}$\\(J)\end{tabular} & Reduction   \\
    \midrule
    12-bit Baseline & \num{3.43} & \num{3.4} & $1\times$\\
    No Adversaries & \num{3.93} & \num{3.87} & $1.14\times$\\
    99\% Adversaries & \num{0.65} & \num{0.61} & $0.18\times$\\
    1\% Adversaries & \num{3.9} & \num{3.86} & $1.13\times$\\
    \bottomrule
  \end{tabular}
  }
\end{table}
\section{Conclusion}
In this paper, we propose ANS, a novel method for identifying CNN layers that are more sensitive to adversarial attacks based on their intermediate activation features. We demonstrate the value of ANS by using it to develop a detection-based method for identifying adversarial examples. In this approach, we append a binary classifier, trained on activations, to a layer that is more vulnerable to adversarial attacks. We show that our method improves state-of-the-art results on the CIFAR-10 and CIFAR-100 datasets. To facilitate efficient computing, we analyze the effects of quantization on our detector-augmented network to find the optimal bit-widths for the CNN and detector. Additionally, we perform hardware energy analysis on this quantized model using a precision scalable hardware accelerator to demonstrate that our method not only improves previous methods, but also is more compute efficient than standalone network architectures. We propose ANS-based detectors as a robust and efficient method for preventing adversarial attacks, and we encourage future research to continue analyzing adversarial examples from a structural perspective to help advance the development of safe artificial intelligence.

\section{Acknowledgement}
This work was supported in part by C-BRIC, Center for Brain-inspired Computing, a JUMP center sponsored by DARPA and SRC, the National Science Foundation (Grant\#1947826), the Technology Innovation Institute, Abu Dhabi and the Amazon Research Award.


\bibliography{./bibliography/IEEEabrv,./bibliography/bibfile.bib}

\small
\end{document}